\documentclass[a4paper,fleqn]{cas-dc}

\usepackage[authoryear,longnamesfirst]{natbib}

\def\tsc#1{\csdef{#1}{\textsc{\lowercase{#1}}\xspace}}
\tsc{WGM}
\tsc{QE}
\usepackage{amsthm}
\newdefinition{rmk}{Remark}
\usepackage{algorithm}
\usepackage{algpseudocode}
\usepackage{amssymb}
\usepackage{subcaption} 
\usepackage{graphicx}

\begin{document}
\let\WriteBookmarks\relax
\def\floatpagepagefraction{1}
\def\textpagefraction{.001}

\shorttitle{Time-MAGNET}    

\shortauthors{Bandyopadhyay et~al.}  

\title [mode = title]{Time-MAGNET: Federated Multimodal Graph Neural Expert Transformer for Human Activity Recognition}  



\author[1]{Asmit Bandyopadhyay}[orcid=0009-0007-7260-0979]
\fnmark[1]
\ead{asmitbandyopadhyay08@gmail.com}
\credit{Conceptualization, Methodology, Writing - Original Draft, Software, Visualization}
\affiliation[1]{organization={Center for Computational Natural Sciences and Bioinformatics, International Institute of Information Technology},
            addressline={},
            city={Hyderabad},
            postcode={500032},
            state={Telengana},
            country={India}}

\author[2]{Tanmay Sen}
\fnmark[1]
\ead{sentanmay518@gmail.com}
\credit{Data Curation, Validation, Writing - Review \& Editing, Supervision}
\affiliation[2]{organization={SQC \& OR Unit, Indian Statistical Institute},
            city={Kolkata},
            postcode={700108},
            state={West Bengal},
            country={India}}

\author[3]{Rohit Basu}
\ead{rohitbasu8953@gmail.com}
\credit{Software}
\affiliation[3]{organization={SQC \& OR Unit, Indian Statistical Institute},
            city={Bangalore},
            postcode={560059},
            state={West Bengal},
            country={India}}

\author[4]{Swagatam Das}
\ead{swagatamdas19@yahoo.co.in}
\credit{Supervision, Writing - Review \& Editing}
\affiliation[3]{organization={Electronics and Communication Sciences Unit, Indian Statistical Institute},
            city={Kolkata},
            postcode={700108},
            state={West Bengal},
            country={India}}

% % Corresponding author text
% \cortext[1]{Corresponding author: Asmit Bandyopadhyay}
% \cortext[2]{Corresponding author: Tanmay Sen}

\fntext[1]{Corresponding author}

% Links
\nonumnote{Code available at: \url{https://github.com/Rivu04/Time-MAGNET}}
\nonumnote{Datasets available at: \url{https://doi.org/10.24432/C59K6T}}

\begin{abstract}
Human Activity Recognition (HAR) plays a vital role in applications such as fitness tracking, smart homes, and healthcare monitoring. Single modality HAR systems rely on a single source of sensory information and can therefore be affected by sensor specific noise and signal degradation, which may limit their ability to distinguish between activities. To address these limitations, this work proposes Time-MAGNET, a novel multimodal learning framework that adaptively fuses heterogeneous sensing modalities to learn discriminative representations for HAR. For multimodal fusion, we propose the Multimodal Adaptive Graph Neural Expert Transformer (MAGNET), which adaptively models interactions among heterogeneous modalities using graph attention and a Mixture-of-Experts architecture. To capture complex temporal dependencies, we employ a lightweight encoder-only T5 architecture customized for multimodal HAR and efficiently trained using Low-Rank Adaptation (LoRA).

In real world applications, multimodal HAR data are typically distributed across devices or users and cannot always be centrally collected due to privacy, data ownership, and communication constraints. This motivates a federated formulation of multimodal HAR, where clients collaboratively learn a shared model without exchanging raw multimodal data. However, the distributed nature of HAR introduces statistical and system heterogeneity across clients, which can challenge conventional federated aggregation methods. To address this issue, we propose Federated Proximal Adaptive Weighting  (\textit{FedPAW}), a novel federated aggregation algorithm for stable and effective model aggregation under heterogeneous client settings.
Extensive experiments on the MEx dataset show that Time-MAGNET achieves accuracies of 0.9555 and 0.8997 in centralized and federated settings, respectively, outperforming the considered baseline methods. The results indicate that adaptive multimodal fusion and LoRA based temporal modeling improve centralized HAR performance, while FedPAW provides effective aggregation under heterogeneous federated clients.
 
 \end{abstract}

\begin{highlights}

\item We designed a customized encoder-only T5 architecture and trained it from scratch without pretrained weights. LoRA adapters were inserted into the encoder attention layers, and all other randomly initialized parameters were frozen.  Learning was performed entirely through the LoRA parameters.

% \item We develop a lightweight encoder-only Transformer based on the T5 architecture, initialized and trained from scratch without using pretrained weights.

\item For T5 encoder input tokens, we adopt a per time step, channel dependent patching strategy, where each time step's multivariate input is embedded into a fixed dimensional token vector compatible with T5's input dimension.

\item We designed and trained a custom CNN encoder called Dual Attention Residual Temporal Convolutional Neural Network (DART-CNN) for extracting highly discriminative spatio-temporal embeddings from image modalities.

% \item We propose a novel multimodal fusion strategy called MAGNET, which integrates Swiglu as the feed forward network, a Graph Attention Network (GAT) and a Mixture of Experts (MoE) framework to capture complex intermodal dependencies.

\item  We propose a novel multimodal fusion strategy called MAGNET that jointly models intermodal relationships using Graph Attention Network (GAT) while adaptively selecting modality specific experts through a  Mixture of Experts (MoE) framework with a SwiGLU based feed forward network. 
% This design enables more effective fusion of heterogeneous sensor modalities than conventional fixed-fusion approaches

\item We developed a novel federated learning algorithm called Federated Proximal Adaptive Weighting (FedPAW) for robust and efficient federated optimization. FedPAW integrates proximal regularization with adaptive weighting, enabling stable convergence under heterogeneous data distributions and communication constraints. This algorithm advances federated learning by improving both training stability and generalization performance across diverse client environments.

\item To the best of our knowledge, this study introduces the first federated learning framework that leverages a customized T5 model for time series data modality embedding and DART-CNN for image modality embedding, collaboratively training multimodal human activity recognition. This approach effectively mitigates data scarcity issues while ensuring the privacy of client data.
\end{highlights}

%\nocite{*}

% Keywords
\begin{keywords}
 Human Activity Recognition (HAR)\sep Federated Learning\sep Multimodal Fusion \sep Graph Attention Network (GAT) \sep Mixture of Experts (MoE) \sep T5 Architecture \sep LoRA
\end{keywords}

\maketitle

% Main text
% \section{}\label{}
\section{Introduction}\label{intro}
Human Activity Recognition (HAR) is a fundamental problem in sensor based machine learning, where the goal is to detect and classify a person’s physical activities based on data recorded by various sensors. With the technological advancement of mobile and wearable devices such as smartphones, smartwatches, fitness bands, and ambient cameras, activity sensing has become increasingly pervasive. These sensors continuously record diverse types of high frequency data such as linear acceleration, angular velocity, spatial location, and environmental context, thus reflecting user behaviour in real time. HAR plays an essential role in a wide array of applications, including remote patient monitoring, elderly care, rehabilitation support, personalised fitness coaching, and smart home automation.

Each sensor plays a vital role in ensuring a continuous flow of information about the underlying task. But single-sensor (unimodal) data often lack contextual richness and robustness and may lead to misinterpretations (e.g., wrist worn sensors can misinterpret activities like brushing as eating owing to similar hand movements). Addressing the problem using multimodal data leads to improved performance as compared to relying on a single data modality~\citep{shaikh2024cnns}, and hence in recent years, HAR has transitioned from unimodal to multimodal setups, where multiple heterogeneous sensor streams (e.g. accelerometer, depth camera, etc. operate simultaneously) are integrated to enable richer and more accurate activity understanding. Sensor data may be collected under either controlled laboratory conditions or in naturalistic real world settings. Multimodal data sets based on labs such as UCI-HAR~\citep{anguita2013public} and UTD-MHAD~\citep{chen2015utd} depict predefined activity sequences performed under supervision. Whereas, real world data sets such as PAMAP2~\citep{Reiss2012IntroducingAN} and CAPTURE-24~\citep{chan2024capture} capture continuous, unscripted behaviour over long durations, thus reflecting the complexities of real life activity recognition. However, effectively modelling this huge volume of complex data is often challenging due to varying sampling rates, noise levels, and modality specific traits. Parallelly, recent developments in Large Language Models (LLMs) have shown their capacity to model complex sequential structures and long range dependencies~\citep{jin2023time} as well as short term transitions, which is crucial for recognizing overlapping or ambiguous activities. Despite being originally designed for text data, these can be adapted to time series tasks by encoding temporal information as sequences. Their self-attention mechanism~\citep{vaswani2017attention} enables them to focus on important time steps, making them powerful tools for processing complex multimodal time series data. 

HAR tasks often need to integrate diverse sensor modalities into a unified representation for capturing richer and more accurate insights. Traditional fusion methods, such as simple concatenation, often miss the nuanced relationships between these modalities. Graph Attention Network (GAT)~\citep{velivckovic2017graph} offers a more adaptive solution by modelling sensors as nodes in a graph and learning attention based connections, helping the system focus on the most relevant signals for a given context. Building on this, the Mixture of Experts (MoE)~\citep{shazeer2017} introduces multiple specialized subnetworks, where a gating mechanism routes each input to the most appropriate experts and thereby makes the model more responsive to diverse user behaviours. 

Moreover, human activity data are often very sensitive, so it is necessary to keep them private. But at the same time, the data is required to train the models. Hence, Federated Learning (FL)~\citep{mcmahan2017communication} has become more applicable in this domain. It enables privacy preserving training directly on edge devices by allowing models to learn collaboratively without ever sharing raw data with the central server. This is particularly important for deploying HAR solutions in scalable, energy efficient wearables and IoT based systems.

Despite the promise of decentralized learning, real-world HAR environments present several challenges that must be addressed simultaneously. First, system heterogeneity arises from differences in the computational resources, battery capacity, and network connectivity of participating devices. Second, statistical heterogeneity arises from the non-IID distribution of activity patterns across users, which may result from differences in individual behaviors and sensing conditions. Third, local training can introduce variability in model updates due to differences in local data, optimization trajectories, and stochastic training, potentially leading to unstable global convergence.
% \begin{itemize}
% (i). \textbf{System Heterogeneity:} Varied computational capabilities, battery levels, and network bandwidth across diverse edge devices. (ii). \textbf{Statistical Heterogeneity:} Non-identical and independent distribution (non-IID) of activity patterns resulting from individual user behaviors and physiological differences. (iii). \textbf{Convergence Instability:} Stochastic noise and erratic local updates that hinder global model consistency and synchronization.
% \end{itemize}

Current state-of-the-art algorithms typically address these challenges in isolation; for instance, FedProx targets statistical drift while FedNova addresses system stragglers. In practical  scenarios, however, these conditions frequently occur in tandem. To address this gap, we propose a novel federated learning algorithm termed \textbf{FedPAW} (\textit{Federated Proximal Adaptive Weighting}). FedPAW uniquely unifies these three properties into a single, robust framework designed to navigate the compounded challenges of real-world HAR deployments.

% Furthermore, while FL ensures data locality, it does not inherently prevent information leakage from shared model gradients. To fortify the framework against such risks, \textit{Differential Privacy} mechanisms are integrated by adding carefully calibrated noise to the training process \citep{10.1007/11787006_1}. This ensures that the inclusion or exclusion of any individual’s data point does not significantly alter the system's behavior, a property mathematically quantified by the privacy budget $(\epsilon, \delta)$ \citep{10.1007/11787006_1}. The integration of DP with the FedPAIN framework creates a comprehensive privacy-preserving architecture that addresses both data locality and individual record privacy \citep{10.1145/3682068}.

% In this paper, we have tried to integrate the aforementioned techniques and devise an architecture for robust activity recognition. 

This work introduces a new learning paradigm for human activity recognition based on a customized encoder-only T5 architecture, where model learning is driven exclusively through low-rank adaptations. The main contributions of our paper are as follows:
\begin{itemize}
\item We designed a customized encoder only T5 architecture trained entirely from scratch, without pretrained weights. To achieve parameter efficient learning, we integrated LoRA adapters into the T5 encoder's attention mechanism and froze all other parameters after random weight initialization so that only the LoRA parameters were trained. This design forces the network to learn exclusively through low rank adaptations, providing a novel perspective on the representational capacity of LoRA when used not for fine tuning but as the sole trainable component in full scale training. 

\item For T5 encoder input tokens, we adopt a per time step, channel dependent patching strategy, where each time step's multivariate input is embedded into a fixed dimensional token vector compatible with T5's input dimension.

\item We designed and trained a custom CNN encoder called Dual Attention Residual Temporal Convolutional Neural Network (DART-CNN) for extracting highly discriminative spatio-temporal embeddings from image modalities.

% \item We propose a novel multimodal fusion strategy called MAGNET, which integrates Swiglu as the feed forward network, a Graph Attention Network (GAT) and a Mixture of Experts (MoE) framework to capture complex intermodal dependencies.

\item We propose MAGNET, a novel multimodal fusion paradigm that captures complex cross-modal dependencies by jointly modeling relational interactions through Graph Attention Networks (GAT), adaptive expert specialization via a Mixture-of-Experts architecture, and expressive nonlinear transformations using SwiGLU feed-forward networks.

\item We developed a novel federated learning algorithm called Federated Proximal Adaptive Weighting  (FedPAW) for robust and efficient federated optimization. FedPAW integrates proximal regularization with adaptive weighting, enabling stable convergence under heterogeneous data distributions and communication constraints. This algorithm advances federated learning by improving both training stability and generalization performance across diverse client environments.

% \item We propose  Federated Proximal Adaptive INertial Aggregation (FedPAIN), a novel federated optimization algorithm that establishes an adaptive proximal–inertial optimization framework for heterogeneous federated learning. By jointly exploiting proximal regularization and adaptive inertial aggregation, FedPAIN achieves stable convergence, communication-efficient optimization, and improved generalization across diverse client populations.

% \item To the best of our knowledge, this study introduces the first differentially private federated learning framework that leverages a customized T5 model for time series data modality embedding and DART-CNN for image modality embedding, collaboratively training multimodal human activity recognition. This approach effectively mitigates data scarcity issues while ensuring the privacy of client data.

\item To the best of our knowledge, this is the first federated learning framework for multimodal human activity recognition that jointly employs a customized T5 architecture for time-series modality embedding and DART-CNN for image modality embedding. The proposed framework enables collaborative learning while alleviating the impact of limited local training data.
\end{itemize}

\section{Related Works}\label{reworks}
Initially, HAR tasks were carried out by traditional machine learning classifiers like KNN, SVM, Gaussian Mixture Models, and Random Forest\citep{attal2015physical}. These models were found to perform well on clean, well segmented data but were susceptible to real world irregularities like sensor noise, user variability, and device placement\citep{gil2023reducing}. 

Beyond the choice of classifier, early work on wearable HAR also examined how sensor type, body placement, and data collection conditions affect recognition. Surveys of wearable sensor HAR discuss the range of sensing modalities and the challenges of deploying activity recognition in everyday settings\citep{6365160,10.1145/2499621}. Benchmark efforts such as the OPPORTUNITY challenge further provided a common basis for evaluating recognition from multiple body worn sensors, including in settings with varied activities and sensor configurations\citep{chavarriaga2013opportunity}. These studies highlight that performance on a single modality dataset may not reflect how well a model generalizes to new users, placements or environments.

Multimodal sensor data fusion was also limited to simple and less efficient feature, concatenation\citep{zhao2024deep}. The transition to deep learning enabled automatic feature learning from raw input, free from the constraints imposed by manual engineering. CNNs and RNNs (e.g., LSTMs, GRUs) came out to be de facto tools, with hybrid models such as DeepConvLSTM\citep{ordonez2016deep} and bidirectional LSTMs\citep{hammerla2016deep}, producing phenomenal results. Yet, they couldn’t capture long range dependencies due to architectural limitations such as restricted receptive fields, vanishing gradients, and scalability\citep{vaswani2017attention}. This led to the development of Transformer based models with an aim to capture long range spatial temporal patterns and cross modal interactions by leveraging the attention mechanism, as performed in\citep{ahn2023star}. TEHAR\citep{mahmud2020human}, TTN\citep{xiao2022two}, and HART\citep{ek2023transformer} are other notable works in this regard, particularly addressing issues with data heterogeneity and improved generalization. 

In addition to hybrid temporal architectures, CNN based models have also been designed to learn from several sensor modalities jointly. For example, multimodal convolutional networks combine accelerometer and gyroscope signals within a shared deep learning framework for activity recognition\citep{ha2015multi}. Such approaches reduce reliance on separately engineered features, while the learned representation can exploit complementary information across sensor channels.

Going a step further from Transformers, LLMs are now being investigated for HAR tasks. A major contribution is HARGPT\citep{ji2024hargpt}, which suggests that LLMs, when well aligned and prompted, can perform activity recognition effectively even in zero shot setups. But sophisticated models such as LLMs and Transformers are difficult to deploy in centralized configurations, particularly concerning data privacy and computation load\citep{nguyen2022federated}. 

Federated Learning (FL)\citep{mcmahan2017communication} comes in to remedy this situation by allowing distributed model training on devices without sharing raw data with the central server. Early work by Sozinov et al. examined federated learning for human activity recognition and described a workflow for training activity recognition models across clients\citep{sozinov2018human}. Their study also highlighted communication cost as a practical consideration, motivating the use of relatively simple models in resource constrained settings. Tu et al. proposed FedDL, which uses dynamic layer sharing to vary which model layers are shared across clients, with the aim of improving communication efficiency while accounting for differences between clients\citep{tu2021feddl}. They evaluated the approach using multiple sensing modalities, including inertial sensors, ultra wideband, and depth camera data.

A central challenge in federated HAR is that clients may differ in their sensor data and activity distributions, so a single fixed training configuration may not perform consistently across participants. Li et al. studied this issue using a hierarchical clustering based federated approach for HAR, organizing clients to address variation in their data and examining robustness and fairness alongside recognition performance\citep{li2023hierarchical}. These works show that practical federated HAR needs to consider not only how model updates are aggregated, but also how client heterogeneity and communication requirements affect training and evaluation.

Efficient tuning techniques such as LoRA have been integrated with FL to further reduce computational costs by training HAR models in low resource environments\citep{qi2024fdlora,abdel2024federated}. LoRA reduces the number of trainable parameters by learning low rank updates to a model’s weights, providing a parameter efficient approach to adapting large pretrained models\citep{hu2022lora}. Another unique instance, GraFeHTy\citep{9680185}, uses federated Graph Convolutional Networks to classify noisy or unlabeled data while protecting user privacy.

\section{Methodology}

\begin{figure*}
    \centering
    \includegraphics[width=\linewidth,height=8cm]{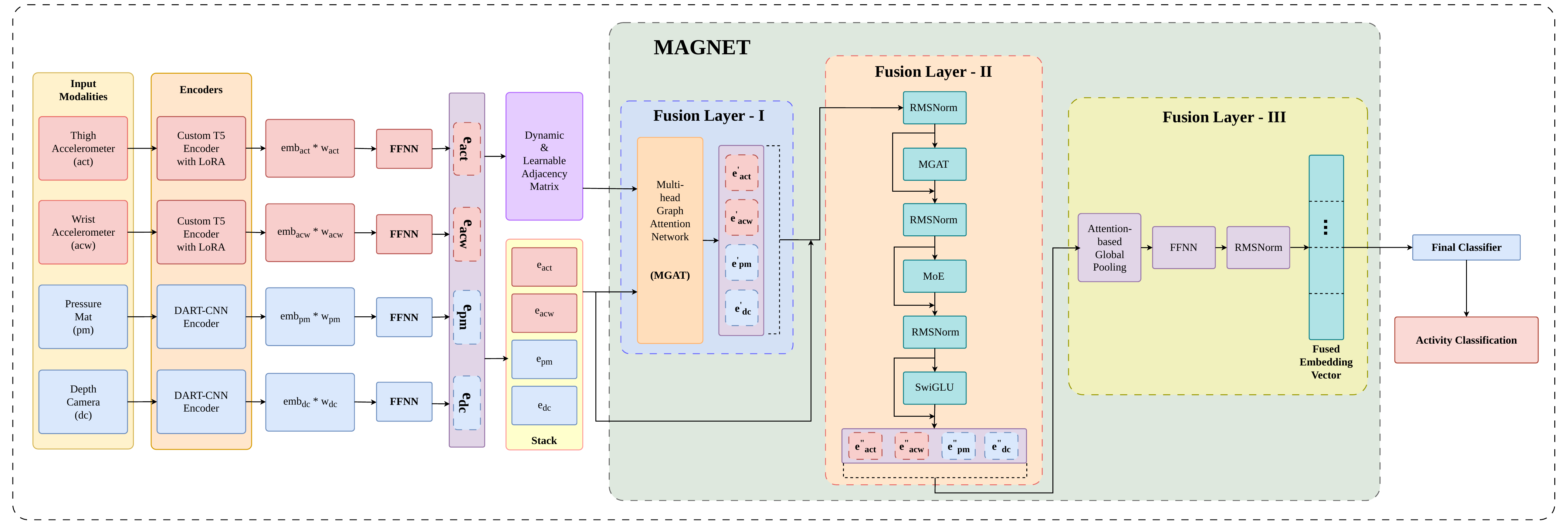} 
    \caption{The proposed HAR architecture - Time-MAGNET.}
    \label{fig1} 
\end{figure*}

This section outlines the proposed multimodal time series model for Federated Human Activity Recognition. The proposed Time-MAGNET framework is a novel multimodal fusion architecture that leverages heterogeneous sensor and image data. This architecture consists of five primary components: (i) a LoRA adapted customized T5 encoder only transformer with learnable positional encodings to encode sequential sensor data, (ii) a Dual Attention Residual Temporal Convolutional Neural Network (DART-CNN) encoder equipped with spatial and channel attention mechanisms for spatial-temporal image encoding, (iii) a Multimodal Adaptive Graph Neural Expert Transformer (MAGNET) for dynamic cross modal feature fusion, and (iv) a new federated learning algorithm, FedPAW, which mitigates statistical and system heterogeneity for stable global convergence. The following subsections describe each component of the architecture in detail. The proposed Time-MAGNET framework is shown in Figure~\ref{fig1}.

\subsection{Time Series Transformer}
Time Series transformers \citep{zerveas2021transformer} utilize transformer architectures to capture patterns in sequential high frequency sensor data. Transformer architectures were originally developed for NLP tasks. These models operate on sequences of discrete tokens, which poses a challenge when dealing with continuous multivariate time series data. To make the data compatible with transformer encoders, the time series is segmented into fixed length chunks referred to as patches. Each patch is then treated as a discrete input token, enabling the transformer to learn temporal dependencies effectively. In this work, we customize the encoder part of the T5 architecture for embedding multivariate time series data. A detailed explanation of the method is as follows:

\paragraph{\textbf{Patching:}} We adopt a channel dependent patching strategy based on individual time steps, as opposed to the channel independent temporal patching used in the PatchTST architecture~\citep{nie2022time}, which aggregates a sequence of consecutive time steps into subseries level patches. Let a multivariate time series be represented as: $\mathbf{X} \in \mathbb{R}^{T \times d}$  with $T$: total number of time steps under consideration, and $d$: number of variables. The objective is to learn representations from the past context $\mathbf{X}_{t-C-1:t} = [\mathbf{x}_{t-C-1}, \cdots, \mathbf{x}_{t}] \in \mathbb{R}^{C \times d}$, where $C$ denotes the context window length. At each time  window, \(\mathbf{X}_{t-C-1:t}\) is the corresponding input segment, where each $\mathbf{x}_i \in \mathbb{R}^{d}$ represents a $d$-dimensional patch.

\paragraph{\textbf{Patch Embedding:}} Each patch $\mathbf{x}_i$, where $i = (t-C-1), \cdots, t$, is mapped to a token vector in $\mathbb{R}^{1 \times D}$,
\begin{eqnarray*}
    z_i = \mathrm{Embed}\left(\mathbf{x}_i\right), \quad z_i \in \mathbb{R}^{1 \times D}. 
\end{eqnarray*}
The resulting patch embeddings, denoted by \(z_i \), serve as the input to the multi-head self-attention mechanism of the Transformer encoder. The \texttt{Embed} operation can be a learnable linear projection, a convolutional encoder, or a MLP. To retain temporal ordering, sinusoidal positional encoding $pos_{i} \in \mathbb{R}^{1 \times D}$ is added with the input embedding vector as:
\begin{eqnarray*}
    \tilde{z}_{i} = z_{i} + pos_{i}.
\end{eqnarray*}

\paragraph{\textbf{Training with LoRA: }} Let $\tilde{Z} = (\tilde{z_1}, \cdots,\tilde{z_C}) \in \mathbb{R}^{C \times D}$ be the input sequence to a multi-head attention layer. For each attention head, the query $Q$, key $K$, and Value matrix $\mathbf{V}$ are derived from $\tilde{Z}$ using respective projection matrices
    \(Q=\tilde{Z}W_Q\), \(K=\tilde{Z}W_K\), \(V=\tilde{Z}W_V\),
where \(W_Q, W_K, W_V \in \mathbb{R}^{D \times d_k}\)\citep{vaswani2017attention} are trainable weight matrices and \(d_k\) is the projection dimension of each of the \(h\) attention heads.
With LoRA, each projection matrix $W_X$ ($X \in \{Q, K, V\}$) is decomposed into a randomly initialized base weight $W_{X,0}$ and a trainable low-rank adaptation  $W_X = W_{X,0} + B_X A_X$,  where  $W_{X,0} \in \mathbb{R}^{D \times d_k}$ are the frozen, randomly initialized base weight matrices, $A_X \in \mathbb{R}^{D \times r}$ and $B_X \in \mathbb{R}^{r \times d_k}$ are the trainable LoRA matrices. 
This formulation significantly reduces computational burden by optimizing only a small subset of the network's parameters. Crucially, \textit{it demonstrates that an entire encoder architecture can be effectively trained using frozen, randomly initialized base weights coupled with these low-rank adaptations}. Consequently, this approach highlights the robust representational capacity of LoRA as a standalone optimization strategy for full scale training from scratch, extending its utility far beyond conventional fine-tuning paradigms without requiring end-to-end parameter updates.

The last hidden state of the T5 encoder's output is then passed through an average pooling layer and a max pooling layer. The concatenated representation of these two layers is then passed through a feedforward neural network (FFNN) to produce the final embedding ${emb}_{T5} \in \mathbb{R}^{B \times d_{model}}$.

\subsection{Dual Attention Residual Temporal Convolutional Neural
Network (DART-CNN) }

The DART-CNN processes 2-dimensional image representations through a series of convolutional blocks with progressively increasing channel dimensions. Given an input tensor \[\mathbf{X} \in \mathbb{R}^{B \times T \times H \times W},\] where B: batch size, T: number of time steps, and $(H\times W)$: spatial dimension of each image, the input is first reshaped to integrate the batch and time dimensions, producing $\mathbf{X}' \in \mathbb{R}^{(B.T) \times 1 \times H \times W}$.  Then each convolutional layer follows the pattern 
\[F_l = \text{GeLU}(\text{BN}_l(\text{Conv}^{C_{l - 1} \rightarrow C_{l}}(F_{l-1}))),\]  
where $\text{C}_l$  the channel progression, $\text{BN}_l$  batch normalization and the convolutional layers $F_l$ extract hierarchical spatial features from the input image data. 

Following feature extraction from the convolutional layers, we applied a dual attention mechanism, consisting of spatial and channel attention, based on the CBAM architecture \citep{10.1007/978-3-030-01234-2_1}, to improve feature representation. Spatial attention generates a spatial weight map, highlighting important locations across the feature map, whereas channel attention \citep{8578843} focuses on the interdependencies between feature channels. Spatial attention weights and channel attention weights are computed as:
\begin{equation*}
    \begin{aligned}
        A_{\text{spatial}} = \sigma\left(\text{Conv}_{1 \times 1}\left(F_{l}\right)\right), \,\,
        \\
        A_{\text{channel}} = \sigma\left(W_2 \cdot \text{ReLU}\left(W_1 \cdot \text{GAP}\left(F_{l}\right)\right)\right),
    \end{aligned}
\end{equation*}
where $\sigma$ is the sigmoid activation function, $\text{GAP}$ denotes global average pooling, $W_1 \in \mathbb{R}^{C/r \times C}$ and $W_2 \in \mathbb{R}^{C \times C/r}$ are learnable parameters with reduction ratio $r$. The final attention weighted feature map can be obtained by 
$$F_{\text{attn}} = F_{l} \odot A_{\text{spatial}} \odot A_{\text{channel}},$$ where $\odot$ denotes element wise multiplication. 

Following the attention mechanism, global average pooling is applied to ${F}_{attn}$ to reduce its spatial dimensions to 1x1, followed by a flattening operation. The resulting feature vector is then passed through a FFNN, which maps it to a fixed embedding dimension $D_{emb}$. This embedding $F_{out} \in \mathbb{R}^{(B\times T) \times D_{emb}}$ is then reshaped back to $\mathbb{R}^{B \times T \times D_{emb}}$ to recover the temporal dimension, preparing the data for sequential modelling.
 
Finally, to capture temporal dynamics within the sequence of features, a stack of recurrent neural networks (LSTM, GRU) with bidirectional layers is applied. The output of each layer serves as the input to the subsequent layer. In the end, the output of these recurrent layers $h_{\text{rec}} = \text{BiGRU}(\text{BiLSTM}({F}_{\text{out}}))$ is combined with a residual connection \citep{7780459} from the fully connected initial output, $h_{\text{rec}} = h_{\text{rec}} + {F}_{\text{out}}$. The final DART-CNN representation is obtained by performing a global temporal mean pooling operation on $h_{\text{rec}}$, resulting in a fixed size representation $emb_{dart} \in \mathbb{R}^{B \times D_{emb}}$.\\

After extracting representations from the T5 encoder and DART-CNN, the model Time-MAGNET incorporates learnable modality specific weights as:
\[\text{emb}_{{weighted}}^{(m)} = w_m \times \text{emb}_m,\] 
where, \(\text{emb} = \{\text{emb}_{dart}^{dc}, \,\text{emb}_{dart}^{pm}, \,\text{emb}_{T5}^{act}, \,\text{emb}_{T5}^{acw}\}\) and \(w = \{w_{act}, \,w_{acw}, \,w_{dc},\, w_{pm}\}\) are learnable parameters that adaptively weight the contribution of each representation of encoded modality based on the characteristics of the data. These embeddings are then projected through a modality specific FFNN, and the resulting embeddings are passed as input to MAGNET's Fusion Layer-I (see Figure~\ref{fig1}).

\subsection{Multi-modal Adaptive Graph Neural Expert Transformer
(MAGNET)}

MAGNET serves as the core fusion mechanism, integrating multimodal embeddings through a sophisticated combination of GAT and MoE architectures. The fusion addresses the challenge of learning complex inter-modal relationships while maintaining computational efficiency through sparse expert routing. The construction of the adjacency matrix for graph attention employs both dynamic and learnable components to form a directed graph structure.
The dynamic adjacency matrix $A_{dynamic} = (\text{cosine}(E_i, E_j) + 1)/2$ is computed using cosine similarity between normalized embeddings $E_i$ and $E_j$. Meanwhile, the learnable parameters $A_{learn} = \sigma(W_{adj})$, where $W_{adj} \in \mathbb{R}^{M \times M}$ denotes the trainable inter-modal logits and $M$ represents the number of modalities, allow the model to discover task-specific modal relationships during training. The final adjacency matrix is computed as follows: 
\begin{center}
    $A_{final} = A_{dynamic} \odot A_{learn} + 0.5 \cdot I$,
\end{center}
where $I$ is the identity matrix for self-loops. Crucially, because the learnable weight matrix $W_{adj}$ is not constrained to be symmetric, $A_{final}$ operates as an asymmetric adjacency matrix. This formulates the multimodal interactions as a directed graph, enabling the network to learn distinct directional dependencies (i.e., the optimal information flow from modality $i$ to modality $j$ may differ from $j$ to $i$).
The final representations of each node $h_i$ using multi-head graph attention (MGAT) \citep{velivckovic2017graph} are computed as: 
\begin{center}
    $h_i = \sigma(\sum_{j} \alpha_{ij}' W h_j + b_i),$
\end{center}
with $\alpha_{ij}' = \alpha_{ij} A_{final}$ denoting the weighted attention score, $W$ the learnable weight matrix, and bias terms $b_i$.\\

In Fusion Layer-II, the graph attention sublayer processes inter-modal relationships, followed by a MoE module that routes information through specialized expert networks. The sparse MoE \citep{shazeer2017} implementation uses top-k routing with entropy based load balancing to ensure uniform expert utilization. Each expert network employs skip connections between hidden layers, normalized using RMSNorm \citep{zhang2019root} for enhanced training stability. The final sublayer applies SwiGLU as a feed forward neural network \citep{shazeer2020glu}, combining the benefits of Swish activation with gated linear units. It is noted that load balancing in the MoE is achieved through an entropy regularization term that encourages uniform usage of experts across the batch. \\

The Fusion Layer-III process concludes with a global attention weighted mean pooling, where modality importance weights are computed as $w_i = \text{softmax}(\text{mean}(h_i))$ from the averaged hidden states. The final fused representation is obtained as $h_{fused} = \sum_{i=1}^M w_i \cdot h_i$, where $h_i$ is the output of Fusion Layer II. This undergoes projection through a FFN and RMSNorm, producing a single, comprehensive feature vector that captures both individual modality characteristics and cross modal interactions essential for activity classification.

MAGNET is the multimodal fusion module of Time-MAGNET framework. Its objective is to model modality interactions while allowing the strength and direction of these interactions to adapt to both the observed modality representations and the learning task. Rather than treating the modality relationships as fixed or assuming symmetric interactions, MAGNET constructs an adaptive directed modality graph and subsequently applies modality-dependent expert transformations.

Let $\mathbf{E}_i\in\mathbb{R}^{d}$ denote the representation of modality $i$, where $i\in{1,\ldots,M}$. We first construct a representation-dependent adjacency component from the pairwise cosine similarity between normalized modality representations:

$$
A^{\mathrm{dyn}}_{ij}
=
\frac{\cos(\mathbf{E}_i,\mathbf{E}_j)+1}{2}.
$$

This component captures the instantaneous similarity between modality representations. Since cosine similarity is symmetric, it alone cannot represent directional dependencies. We therefore introduce a learnable inter-modal relation matrix
$A^{\mathrm{learn}}
=
\sigma(W_{\mathrm{adj}}),
\,\,
W_{\mathrm{adj}}\in\mathbb{R}^{M\times M}$,
where $W_{\mathrm{adj}}$ is not constrained to be symmetric. The final edge-weight matrix is defined as

\[
A^{\mathrm{final}}
=
A^{\mathrm{dyn}}\odot A^{\mathrm{learn}}
+
\lambda I,
\]

\noindent where $\odot$ denotes element-wise multiplication, $I$ provides self-connections, and $\lambda=0.5$ in our implementation.\\

This formulation combines representation dependent similarity with task-learned inter-modal relations. In particular, $A^{\mathrm{final}}*{ij}$ and $A^{\mathrm{final}}*{ji}$ are not required to be equal, allowing the model to represent asymmetric information flow between modalities. Thus, the graph can capture cases in which the contribution of modality $i$ to modality $j$ differs from the contribution of modality $j$ to modality $i$. The resulting modality graph is processed using multi-head graph attention. For node $i$, the graph-attention representation is given by

$$
\mathbf{h}_i
=
\sigma
\left(
\sum_j
\alpha'_{ij}W\mathbf{h}_j
+
\mathbf{b}_i
\right),
$$

where $\alpha'*{ij}=\alpha*{ij}A^{\mathrm{final}}_{ij}$ denotes the adjacency-weighted attention coefficient. The graph-attention operation therefore uses both learned attention coefficients and the adaptive inter-modal edge weights.\\

Following graph-based interaction modeling, Fusion Layer-II employs a sparse Mixture-of-Experts (MoE) module to provide modality dependent feature transformation. A top-$k$ router assigns each representation to a subset of expert networks, allowing different experts to specialize in different patterns of multimodal information. An entropy based load-balancing regularizer is used to discourage excessive concentration of routing decisions on a small number of experts. Each expert employs residual connections and RMSNorm, while the subsequent feed-forward transformation uses SwiGLU.\\

Finally, Fusion Layer-III performs adaptive modality aggregation. For each modality representation $\mathbf{h}_i$, an importance weight is obtained from its aggregated hidden representation:
$
w_i
=
\operatorname{softmax}
\left(
\operatorname{mean}(\mathbf{h}_i)
\right),
$
and the fused representation is computed as
$$
\mathbf{h}_{\mathrm{fused}}
=
\sum_{i=1}^{M}w_i\mathbf{h}_i.
$$

The fused representation is subsequently transformed by a feed-forward projection and RMSNorm before classification. Consequently, MAGNET performs three successive operations: This design allows the fusion process to account for both pairwise inter-modal dependencies and the relative contribution of individual modalities.

\subsection{Missing Modality Handling and Classification}

The architecture is explicitly designed to handle missing sensor modalities during inference. If a modality tensor is omitted (e.g., due to sensor failure), the model substitutes it with a globally learnable missing embedding $\mathbf{m} \in \mathbb{R}^{d_{model}}$. Furthermore, modality specific learnable scalars ($\gamma \in \mathbb{R}^{M}$) are applied prior to fusion, allowing the network to dynamically assign importance or penalize unreliable sensor streams.

To map the comprehensive feature vector (produced by Fusion Layer-III) to the final $C$ activity classes, the representation is passed to a hierarchical multi-layer feed-forward neural network (FFNN). This classification head incorporates progressive dimensionality reduction ($d \to d/2 \to d/4 \to C$), GELU activations, and heavy dropout at each stage to prevent overfitting and ensure robust prediction of activity labels.

\subsection{Federated Learning}
\label{sec:fedpain}

\begin{figure*}
\centering
\includegraphics[width=\linewidth]{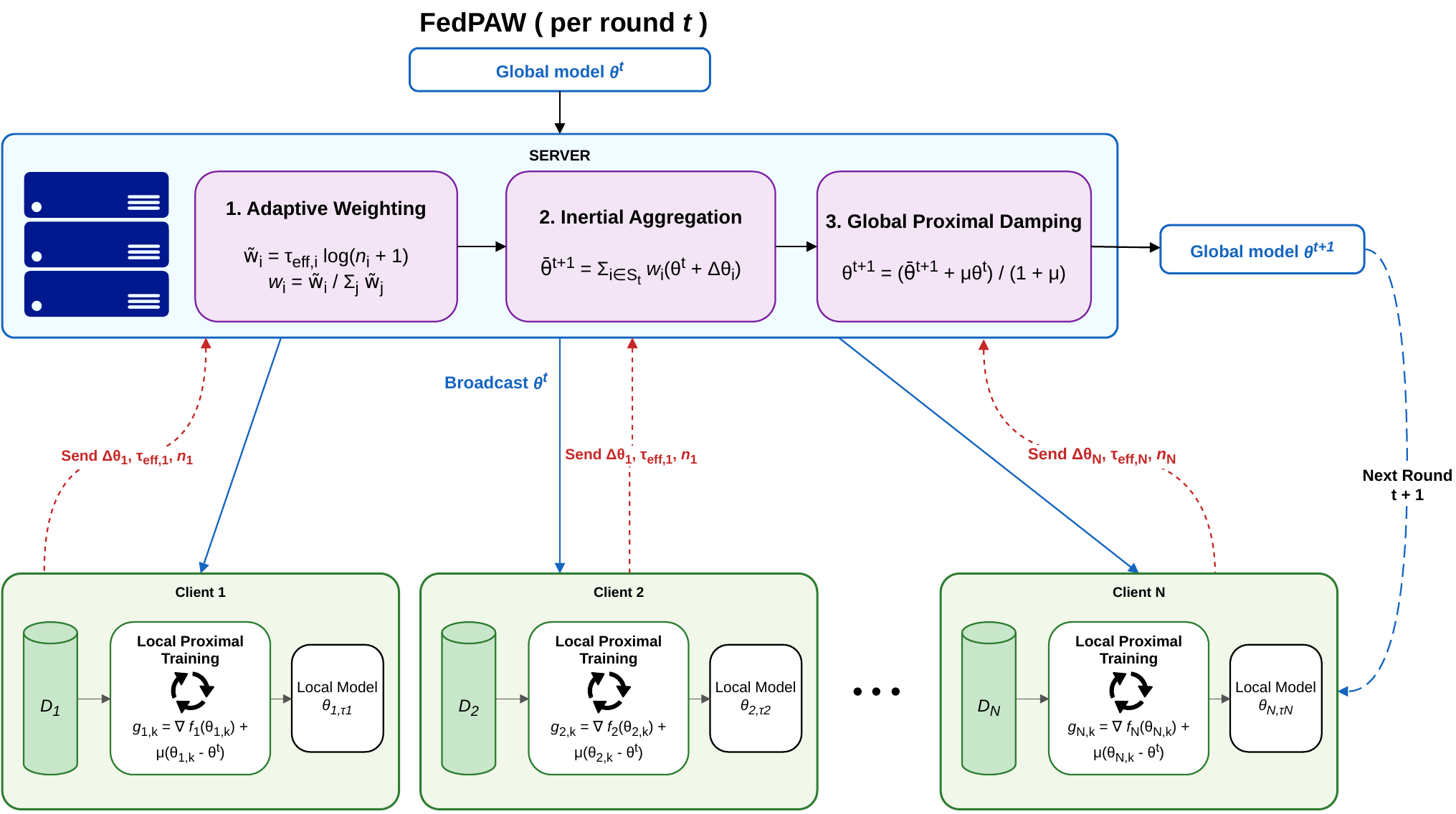}
\caption{The proposed federated learning framework, Federated Proximal Adaptive Weighting (FedPAW).}
\label{fig2}
\end{figure*}

% The multimodal Time-MAGNET  operates on continuous sensor streams collected from individual participants. Directly pooling these recordings at a central server is undesirable because the data are both high-dimensional and potentially sensitive. We therefore train Time-MAGNET in a federated learning (FL) setting~\citep{mcmahan2017communication}, where each participant retains its data locally and only model information is communicated to the server.

Human activity recognition (HAR) relies on sensor data collected from individuals performing different activities under diverse conditions. Centralizing these recordings may raise privacy concerns, particularly when continuous motion and physiological data are involved. Federated learning (FL)~\citep{mcmahan2017communication} enables collaborative model training across participants without transferring their raw sensor data to a central server. Accordingly, we train multimodal Time-MAGNET in a federated setting, where each participant performs local training and shares model updates for global aggregation.

A widely used aggregation strategy in federated learning is Federated Averaging (FedAvg)~\citep{mcmahan2017communication}, which combines locally trained models through a weighted average based on the clients' dataset sizes. However, federated training of multimodal HAR models can involve two forms of heterogeneity. First, participants may have different activity distributions, resulting in variations in their local optimization objectives. Second, differences in dataset sizes and local training conditions may lead to unequal numbers of local updates. Consequently, local training can produce client updates with substantially different magnitudes and directions, which may affect the quality of global aggregation.

Several federated optimization methods address these challenges from different perspectives. FedProx~\citep{MLSYS2020_1f5fe839} introduces a proximal regularization term into the local objective to discourage excessive deviations from the current global model, thereby mitigating client drift. FedNova~\citep{wang2020tackling}, in contrast, normalizes client updates to account for differences in the amount of local computation. Although these methods address complementary aspects of federated optimization, their individual mechanisms do not explicitly account for both challenges within a single optimization procedure.

\paragraph{Federated Proximal Adaptive Weighting (FedPAW):} Motivated by this consideration, we propose \textbf{FedPAW}, a federated optimization method designed to account for client drift and variations in local optimization progress. Unlike conventional aggregation schemes that primarily rely on client dataset sizes, FedPAW assigns adaptive client weights based on the cumulative learning rate applied during local training and a logarithmic transformation of the local sample size. This weighting mechanism accounts for differences in local optimization effort while moderating the influence of clients with larger datasets. During local training, a proximal regularization term penalizes deviations of the local model from the current global model. At the server, an inertial update combines the aggregated client model with the previous global model to regulate the global optimization trajectory. Figure~\ref{fig2} illustrates the overall FedPAW procedure.

\vspace{4pt}

\paragraph{\textbf{Local training.}}
At communication round \(t\), the server randomly selects \(S\) clients
from the \(K\) available clients, forming the participating client set
\(\mathcal{S}_t\subseteq[K]\), with \(|\mathcal{S}_t|=S\).
Each selected client \(i\in\mathcal{S}_t\) initializes its local model as
\(
\theta_{i,0}=\theta^t
\)
and performs local optimization using the following proximal objective:
\begin{equation}
\mathcal{L}_i(\theta;\theta^t)
=
\frac{1}{|\mathcal{B}_i|}
\sum_{(x,y)\in\mathcal{B}_i}
\ell\!\left(f(x;\theta),y\right)
+
\frac{\mu}{2}\|\theta-\theta^t\|_2^2,
\label{eq:local_obj}
\end{equation}
where \(\ell(\cdot,\cdot)\) denotes the class-weighted cross-entropy
loss, \(\mu\) is the proximal coefficient, and \(\mathcal{B}_i\)
denotes a local mini-batch. The proximal term penalizes deviations
from the current global model, thereby discouraging excessive local
updates toward patterns specific to individual participants
\citep{MLSYS2020_1f5fe839}. Because processing the multimodal inputs is memory intensive,
gradients are accumulated over \(A\) consecutive mini-batches before
an AdamW \citep{loshchilov2017decoupled} optimizer update. For a
complete accumulation window, the nominal effective batch size is
\(
B_{\mathrm{eff}}=B\times A,
\)
where \(B\) is the mini-batch size. The learning rate is decayed across local epochs according to
\begin{equation}
\eta_k
=
\eta_0
\gamma^{\left\lfloor (e(k)-1)/s \right\rfloor},
\label{eq:steplr}
\end{equation}
where \(\eta_0\) is the initial learning rate, \(\gamma\) is the
decay factor, \(s\) is the number of local epochs between successive
decays, and \(e(k)\in\{1,\ldots,E\}\) denotes the local epoch
associated with optimizer step \(k\).

The number of optimizer steps alone does not fully reflect the
learning rate schedule applied during local training. In particular,
clients performing different numbers of local updates may accumulate
different learning rate exposures, depending on the number of updates
performed before and after learning-rate decay. To account for this
variation, we record the cumulative learning rate over the local
optimizer updates:
\begin{equation}
\tau_{\mathrm{eff},i}
=
\sum_{k=1}^{T_i}\eta_k,
\label{eq:tau}
\end{equation}
where \(T_i\) denotes the total number of optimizer updates performed
by client \(i\). This quantity provides a learning-rate-based measure
of the client's optimization exposure under the prescribed schedule.
Consequently, updates performed after learning-rate decay contribute
less to \(\tau_{\mathrm{eff},i}\) than those performed earlier in
training.

Each client performs AdamW updates for \(E\) local epochs. Gradient
accumulation triggers an optimizer update after every \(A\) mini-batches,
with any remaining mini-batches at the end of an epoch also contributing
to an optimizer update. If a client performs no optimizer updates, we
set \(\tau_{\mathrm{eff},i}=0\). After completing local optimization,
client \(i\) sends
\(
(\theta_i,\tau_{\mathrm{eff},i},n_i)
\)
to the server, where \(\theta_i\) is its final local model and \(n_i\)
is its local sample count.

\vspace{4pt}

\paragraph{\textbf{Global aggregation:}}
The server aggregates the received local models in three stages.
Unlike FedAvg, where the aggregation weight is proportional to
\(n_i\), FedPAW separates the contribution associated with local
learning rate exposure from the effect of dataset size.

\paragraph{Stage 1: Adaptive sublinear weighting.}

For each participating client \(i\in\mathcal{S}_t\), we define
the raw aggregation weight as
\begin{equation}
{q}_i= \tau_{\mathrm{eff},i}\log(n_i+1).
\label{eq:raw_weights}
\end{equation}

The first factor, \(\tau_{\mathrm{eff},i}\), captures the cumulative
learning rate exposure of the client during local training. The
second factor introduces a sublinear dependence on the local sample
size. In contrast to the linear sample-count weighting used by
FedAvg, the logarithmic transformation moderates the influence of
larger datasets on the aggregation weight. Thus, dataset size
continues to influence each client's contribution without increasing
it proportionally with the number of local samples. 

\noindent The normalized client weights are then obtained as
\begin{equation}
w_i
=
\frac{{q}_i}
{\displaystyle\sum_{j\in\mathcal{S}_t}{q}_j}.
\label{eq:norm_weights}
\end{equation}
If the denominator is smaller than \(10^{-9}\), uniform weights
\(w_i=1/|\mathcal{S}_t|\) are assigned to all participating clients.

\paragraph{Stage 2: Weighted model aggregation.}

The client models are combined according to their normalized
adaptive weights:
\begin{equation}
\bar{\theta}^{t+1}
=
\sum_{i\in\mathcal{S}_t}w_i\theta_i.
\label{eq:weighted_aggregate}
\end{equation}
Since \(\sum_{i\in\mathcal{S}_t}w_i=1\), the aggregate can
equivalently be expressed in terms of the local model displacements
\(\Delta\theta_i=\theta_i-\theta^t\) as
\begin{equation}
\bar{\theta}^{t+1}
=
\theta^t+
\sum_{i\in\mathcal{S}_t}w_i\Delta\theta_i.
\end{equation}
Thus, the server combines the local model displacements relative
to their common initialization \(\theta^t\), with the adaptive
weights determining the contribution of each participating client.

\paragraph{Stage~3: Damped server update.}

The server applies a damped update to combine the weighted client
aggregate with the current global model:
\begin{equation}
\theta^{t+1}
=
\frac{\bar{\theta}^{t+1}+\mu\theta^t}{1+\mu},
\label{eq:global_update}
\end{equation}
where $\mu$ is the same proximal coefficient used in the local training (equation~\ref{eq:local_obj}). Here, it is used to control the degree of server side damping.
Using the displacement representation of the weighted aggregate,
the global update can equivalently be expressed as
\begin{equation}
\theta^{t+1}
=
\theta^t+
\frac{1}{1+\mu}
\sum_{i\in\mathcal{S}_t}w_i\Delta\theta_i.
\end{equation}
Thus, the server applies a fraction \(1/(1+\mu)\) of the
weighted client displacement in each communication round. This
mechanism moderates the influence of the aggregated client updates
and provides additional control over the magnitude of global model
changes.

To ensure numerical stability, each parameter tensor of the updated
global model is checked for non-finite values. If a tensor contains
any \(\mathrm{NaN}\) or \(\mathrm{Inf}\) entries, it is replaced
with the corresponding tensor from the previous global model
\(\theta^t\). The complete FedPAW procedure is outlined in
Algorithm~\ref{alg:fedpaw}.

\begin{algorithm*}[ht]
\caption{FedPAW: Federated Proximal Adaptive Weighting}
\label{alg:fedpaw}
\begin{algorithmic}[1]
\State \textbf{Input:} Initial model $\theta^0$, proximal coefficient and 
server damping coefficient $\mu$, initial learning rate $\eta_0$,
decay factor $\gamma$, decay interval $s$, local epochs $E$,
accumulation steps $A$, clients per round $S$, communication rounds $T$
\For{$t=0,\ldots,T-1$}
    \State Select $S$ clients uniformly at random to form $\mathcal{S}_t$
    \For{each client $i\in\mathcal{S}_t$ \textbf{in parallel}}
        \State $\theta_i\leftarrow\theta^t$; $\tau_{\mathrm{eff},i}\leftarrow0$
        \For{$e=1,\ldots,E$}
            \State $\eta_e\leftarrow\eta_0\gamma^{\lfloor(e-1)/s\rfloor}$
            \State $G_i\leftarrow0$; $a\leftarrow0$
            \For{each mini-batch $\mathcal{B}$ in $\mathcal{D}_i$}
                \State $g_i\leftarrow
                \nabla_{\theta_i}\left[
                \frac{1}{|\mathcal{B}|}
                \sum_{(x,y)\in\mathcal{B}}\ell(f(x;\theta_i),y)
                \right]+\mu(\theta_i-\theta^t)$
                \State $G_i\leftarrow G_i+g_i$; $a\leftarrow a+1$
                \If{$a=A$ or end of epoch}
                    \State $G_i\leftarrow G_i/a$
                    \State Clip $G_i$ and update $\theta_i$ using AdamW
                    with learning rate $\eta_e$
                    \State $\tau_{\mathrm{eff},i}\leftarrow
                    \tau_{\mathrm{eff},i}+\eta_e$
                    \State $G_i\leftarrow0$; $a\leftarrow0$
                \EndIf
            \EndFor
        \EndFor
        \State Send $(\theta_i,\tau_{\mathrm{eff},i},n_i)$ to the server
    \EndFor
    \State $q_i\leftarrow\tau_{\mathrm{eff},i}\log(n_i+1)$
    \hfill $\forall i\in\mathcal{S}_t$
    \If{$\sum_{i\in\mathcal{S}_t} q_i<10^{-9}$}
        \State $w_i\leftarrow1/S$ for all $i\in\mathcal{S}_t$
    \Else
        \State $w_i\leftarrow
        q_i/\sum_{j\in\mathcal{S}_t} q_j$
        \hfill $\forall i\in\mathcal{S}_t$
    \EndIf
    \State $\bar{\theta}^{t+1}\leftarrow
    \sum_{i\in\mathcal{S}_t}w_i\theta_i$
    \State $\theta^{t+1}\leftarrow
    (\bar{\theta}^{t+1}+\mu\theta^t)/(1+\mu)$
    %\State Replace each non-finite parameter tensor in $\theta^{t+1}$
    with its counterpart in $\theta^t$
    \State Revert any tensor in $\theta^{t+1}$ containing $\mathrm{NaN}$/$\mathrm{Inf}$
           to its value in $\theta^t$
\EndFor
\State \Return $\theta^T$
\end{algorithmic}
\end{algorithm*}

\section{Experiment}

\paragraph{Dataset Description:}
To evaluate the effectiveness of the proposed framework, we conduct experiments on the publicly available MEx~\citep{wijekoon2019mex} multimodal dataset. The dataset comprises sensor readings at about 100 Hz from two accelerometers placed on the wrist and thigh of the performer, a pressure mat and a depth camera. There are 30 performers, each performing 7 different physiotherapy exercises: Knee-rolling (class 0), Bridging (class 1), Pelvic tilt (class 2), The Clam (class 3), Repeated Extension in Lying (class 4), Prone punches (class 5), and Superman (class 6), with each exercise lasting up to 60 seconds. The dc (depth cam) modality captures a \(12\times16\) (scaled down from \(240\times320\) to \(12\times16\) using the OpenCV resize) grid of depth measurements at about 15 Hz, providing spatial information about participant positioning, while the pm (pressure mat) modality records a \(32\times16\) grid of pressure values at about 15 Hz, encoding spatial pressure distributions.

\paragraph{Data Preprocessing:}
Timestamps were standardized to Unix epoch format with second precision. Each modality's sampling rate was normalized through linear interpolation: accelerometer data (ACT/ACW) maintained 100 Hz sampling, while depth camera (DC) and pressure mat (PM) data were standardized to 15 Hz. Time series normalization employed z-score standardization per modality.

Sliding window segmentation extracted fixed duration segments with 5 second windows and 1 second increments, providing 80\% overlap between consecutive windows. This approach generated 500 temporal frames for accelerometer data and 75 frames for DC/PM modalities per window. Data augmentation incorporated additive Gaussian noise with zero mean and 0.01 standard deviation, applied exclusively to the accelerometer data segments $\mathbf{x}_{\text{aug}} = \mathbf{x} + \mathcal{N}(0, 0.01^2)$ during dataset initialization.  

\paragraph{Experimental Setup:}
Our experiments were conducted using Python 3.12.12 with PyTorch 2.8.0+cu126 as the primary deep learning framework. Additional dependencies included scikit-learn 1.6.1 for preprocessing and evaluation metrics, pandas 2.2.2 and NumPy 2.0.2 for data manipulation, and the transformers 4.57.1 library for T5 architecture implementation. All experiments were performed on a Kaggle Notebook with an NVIDIA Tesla P100-PCIE-16GB GPU (15.9 GB VRAM) and 31.3 GB system RAM. The CUDA 12.6 runtime environment is used for GPU acceleration. All experiments used multiple random seeds with deterministic CUDA operations to ensure reproducibility.

\subsection{Model Specifications }
\vspace{0.3cm}
\paragraph{Customized T5:} The T5 encoder used a model dimension of 512, comprising a single Transformer encoder layer with 8 attention heads. Each attention head operates with key/value dimensions of 64. The feedforward sublayer is widened with a dimension of 2048, leveraging a gated GELU projection mechanism. A dropout rate of 0.1 is applied to both the attention and feedforward subcomponents, with numerical stability ensured via layer norm $\epsilon = 10^{-6}$. Relative positional biases are discretized over 32 buckets to efficiently model temporal relationships. The LoRA configuration utilized rank $r=16$ with $\alpha=32$, targeting query, key, and value projection matrices. Positional encodings accommodated maximum sequence lengths of 500 for accelerometer data.
\vspace{0.7mm}

% \paragraph{DART-CNN:} The convolutional encoder progressed through channel dimensions [64, 128, 256, 512] with $3\times3$ kernels and stride 1. Batch normalization and GeLU activations followed each convolutional layer. A max pooling operation is applied after the $2^{nd}$ convolutional layer to reduce the spatial dimensions. The attention modules employed a reduction ratio $r=16$ for computational efficiency. The temporal processing flow included bidirectional LSTM (2 layers, 256 hidden units), and GRU (2 layer, 256 hidden units) with 0.1 dropout rates.

\paragraph{DART-CNN:} The convolutional encoder consists of four convolutional layers with channel dimensions of [64, 128, 256, 512], using \(3\times3\) kernels and a stride of 1. Each convolutional layer is followed by batch normalization and GELU activation. A max-pooling operation is applied after the second convolutional layer to reduce the spatial dimensions. Following convolutional feature extraction, spatial and channel attention modules are employed to enhance the representation of discriminative spatial features and inter-channel dependencies. The attention modules use a reduction ratio of \(r=16\). The resulting feature maps are globally average-pooled and projected through a feedforward network to obtain embeddings of dimension \(D_{\mathrm{emb}}\). For temporal modelling, the embeddings are processed sequentially through a two-layer bidirectional LSTM and a two-layer bidirectional GRU, each with 256 hidden units per direction and a dropout rate of 0.1. A residual connection from the initial projected embeddings is incorporated into the recurrent representation, followed by global temporal average pooling to obtain the final fixed-dimensional DART-CNN representation.

% \paragraph{MAGNET Configuration:} The fusion module operated with 4 modalities, transitioning from an initial pre-fusion multi-head graph attention block (Fusion Layer-I) to the core enhanced fusion module (Fusion Layer-II) via a hard skip connection to preserve the initial modal embeddings. Within Fusion Layer-II, learnable skip connections were sequentially integrated across its three sub-components: the multi-head graph attention (mGAT), Mixture of Experts (MoE), and SwiGLU networks, ensuring stable gradient flow and deep feature retention. The mixture of experts employed 4 experts with top-2 selection, while graph attention utilized 8 heads. Load balancing weight $\lambda$ was set to 0.01. RMSNorm epsilon was configured as $10^{-6}$ for numerical stability.

\paragraph{MAGNET Configuration:} The fusion module operates on four modalities and comprises three successive fusion stages. Fusion Layer-I employs an initial multi-head graph attention block for pre-fusion, with a hard skip connection preserving the initial modality embeddings for subsequent processing. Fusion Layer-II consists of three sequential subcomponents: multi-head graph attention (mGAT), a Mixture-of-Experts (MoE) module, and a SwiGLU feedforward network. Learnable skip connections are incorporated across these subcomponents to facilitate gradient flow and preserve intermediate representations. The MoE module employs four experts with top-2 routing, while the graph attention mechanism utilizes eight attention heads. The load-balancing weight $\lambda$ is set to 0.01, and RMSNorm is employed with an epsilon value of \(10^{-6}\) to improve numerical stability.

% An auxiliary load-balancing coefficient of \(\lambda_{\mathrm{MoE}}=0.01\) is used to regulate expert utilization. RMSNorm is employed with an epsilon value of \(10^{-6}\) to improve numerical stability.

\subsection{Baseline Models}
\vspace{0.3cm}
% \paragraph{Chronos-MAGNET: } A variant of our proposed architecture that replaces our custom-trained T5 encoder blocks with Amazon's pre-trained Chronos-T5-mini foundation encoders \citep{ansari2024chronos}. This baseline evaluates whether leveraging a large-scale, pre-trained time-series language model yields more discriminative representations than our domain-specific T5 formulation when paired with the MAGNET fusion framework.

\paragraph{Chronos-MAGNET:}
Chronos-MAGNET is a variant of the proposed architecture in which the custom-trained T5 encoder blocks are replaced with the pre-trained Chronos-T5-mini model developed by Amazon \citep{ansari2024chronos}. The encoder representations extracted from Chronos-T5-mini are subsequently passed to the MAGNET fusion framework. This baseline investigates whether representations learned through large-scale time-series pre-training provide an advantage over those learned by our domain-specific T5 encoder when used within the same multimodal fusion architecture.

\paragraph{Chronos-T5:}
Chronos-T5 is a foundation model based multimodal baseline built upon the pre-trained Chronos-T5-mini architecture \citep{ansari2024chronos}. In this implementation, raw sensor and spatial inputs from individual modalities are independently projected into modality-specific representations and processed using Chronos-T5 encoders. The resulting modality embeddings are concatenated to form a unified sequence, which is subsequently processed by three sequential Chronos-T5 decoder blocks. This baseline evaluates the effectiveness of a sequence-to-sequence foundation-model-based architecture for multimodal fusion without incorporating explicit graph-based structural inductive biases.

\paragraph{BiLSTM: } A highly robust, purely recurrent multimodal architecture. This baseline employs deep, residual Bidirectional Long Short-Term Memory (BiLSTM) networks equipped with attention pooling to encode both temporal and spatial modalities \citep{graves2005framewise}. Cross-modal integration is achieved through a multi-stage BiLSTM fusion module, serving as a high-capacity sequential modelling baseline against our graph-based mechanisms.

\paragraph{DeepConvLSTM: } A scaled-up, multimodal adaptation of the highly established DeepConvLSTM architecture \citep{ordonez2016deep}. It utilizes a deep stack of 1D convolutional layers to extract fine-grained local features, followed by LSTM layers to capture long-term temporal dependencies within each modality independently. The independent representations are then concatenated and passed through a deep multi-layer perceptron (MLP) for classification.

\paragraph{Late Fusion: } An ablation baseline designed to isolate the impact of our proposed MAGNET fusion module. It retains the exact custom T5 and DART-CNN encoders used in our framework but replaces the Graph Attention and Mixture-of-Experts routing with a standard late fusion approach. The independent modality embeddings are simply concatenated and processed by a deep MLP, demonstrating the consequences of omitting cross-modal interactions before the decision level.

\paragraph{Attention Fusion: } A secondary ablation baseline that substitutes our multi-scale graph attention and expert routing with a standard Multi-Head Self-Attention (Transformer) layer \citep{vaswani2017attention}. By treating the independent modality embeddings as a sequence of discrete tokens, this baseline allows us to empirically verify whether our specialized graph-based topology outperforms generic cross-attention for multimodal HAR fusion.

\subsection{Training}
Model training employed \textit{AdamW} optimizer with initial learning rate $1e^{-4}$, weight decay $1e^{-4}$, and gradient clipping at maximum norm 1.0. The \textit{ReduceLROnPlateau} scheduler reduced the learning rate by a factor of 0.5 with patience of 3 epochs and a minimum learning rate $1e^{-6}$. Gradient accumulation over 6 steps enabled effective batch processing with a batch size of 8, achieving an equivalent batch size of 48. Mixed precision training with automatic loss scaling enhanced computational efficiency while maintaining numerical stability. Class imbalance was addressed through the weighted cross-entropy loss. The total loss incorporated MoE load balancing loss with the weighted cross-entropy loss, $\mathcal{L}_{\text{total}} = \mathcal{L}_{\text{CE}} + 0.01 \times \mathcal{L}_{\text{MoE}}$.

\vspace{0.5mm}

% \paragraph{Centralized Training: }
% The dataset was partitioned into 70\% training (21 participants), 10\% validation (3 participants), and 20\% testing (6 participants). Early stopping with a patience of 6 epochs prevented overfitting, monitoring validation accuracy as the primary metric. Maximum training duration was set to 10 epochs with automatic checkpointing of the best performing model based on validation accuracy.  

\paragraph{Centralized Training:}
The dataset was partitioned at the participant level into 70\% training (21 participants), 10\% validation (3 participants), and 20\% testing (6 participants). Early stopping with a patience of 6 epochs was employed to mitigate overfitting, with validation accuracy used as the primary model-selection metric. The maximum training duration was 10 epochs, and the checkpoint achieving the highest validation accuracy was retained for final evaluation.

\vspace{0.5mm}
% \paragraph{Federated Learning: }
% Similar to the centralized setup, the federated learning environment considers 21 clients for training, 3 for validation, and 6 for testing. Local training employs the same optimization strategy for 5 local epochs per client, while 8 out of 21 clients are randomly sampled for each global round (sampling ratio $\approx 0.38$). The server aggregates model parameters using the FedPAW algorithm, which incorporates a proximal term ($\mu = 0.2$) during local training to anchor updates, and weights client updates by their effective local step sizes alongside an inertial server momentum to combat statistical heterogeneity. Memory optimization includes strategic CPU-GPU transfers, gradient accumulation, and explicit cleanup via cache cleaning and garbage collection. The system supports configurable hyperparameters (10 global rounds, 5 local epochs, batch size 8) with automatic CUDA device placement and comprehensive training progress logging.

\paragraph{Federated Training:}
The federated setting used the same participant-level partition, comprising 21 training clients, 3 validation participants, and 6 test participants. Each participating client performed 5 local training epochs using the same optimization strategy as in centralized training. At each global communication round, 8 of the 21 training clients were randomly selected, corresponding to a client sampling ratio of approximately 0.38. The server aggregated client models using the proposed FedPAW algorithm, which incorporates a proximal regularization term (\(\mu=0.2\)) during local optimization to constrain client updates, together with effective-step-size-weighted aggregation and a damped server update to mitigate the effects of statistical heterogeneity. The federated implementation used 10 global communication rounds. Memory management included gradient accumulation, strategic CPU--GPU transfers, and explicit cache cleanup and garbage collection to reduce GPU memory requirements. Automatic CUDA device placement and training-progress logging were also employed.

\section{Results and Discussion}
% We evaluate Time-MAGNET under both centralized and federated training regimes across five complementary axes: (i) training dynamics and convergence behaviour, (ii) overall predictive performance against seven baselines, (iii) the effect of our proposed FedPAW federated optimization algorithm relative to four established alternatives, (iv) fine grained classification quality via confusion matrices, precision-recall curves, and embedding visualisations, and (v) sensitivity to input modality and model complexity and training cost. Together these results characterise not only how well Time-MAGNET performs, but where its gains originate and what they cost computationally.

We evaluate Time-MAGNET under centralized and federated training regimes in terms of convergence behaviour, predictive performance against seven baseline architectures, and the effectiveness of FedPAW compared with four established federated optimization methods. We also examine class-wise performance, learned embedding representations, modality contributions, and computational cost.

% We also examine class-wise performance, learned embedding representations, modality contributions, and computational cost. These experiments assess the model's performance and the contributions of its key components.

\begin{figure*}[htbp]
  \centering
  % Row 1 — Centralised training: Time-MAGNET vs. baseline architectures
  \begin{subfigure}{0.45\textwidth}
    \centering
    \includegraphics[width=\linewidth]{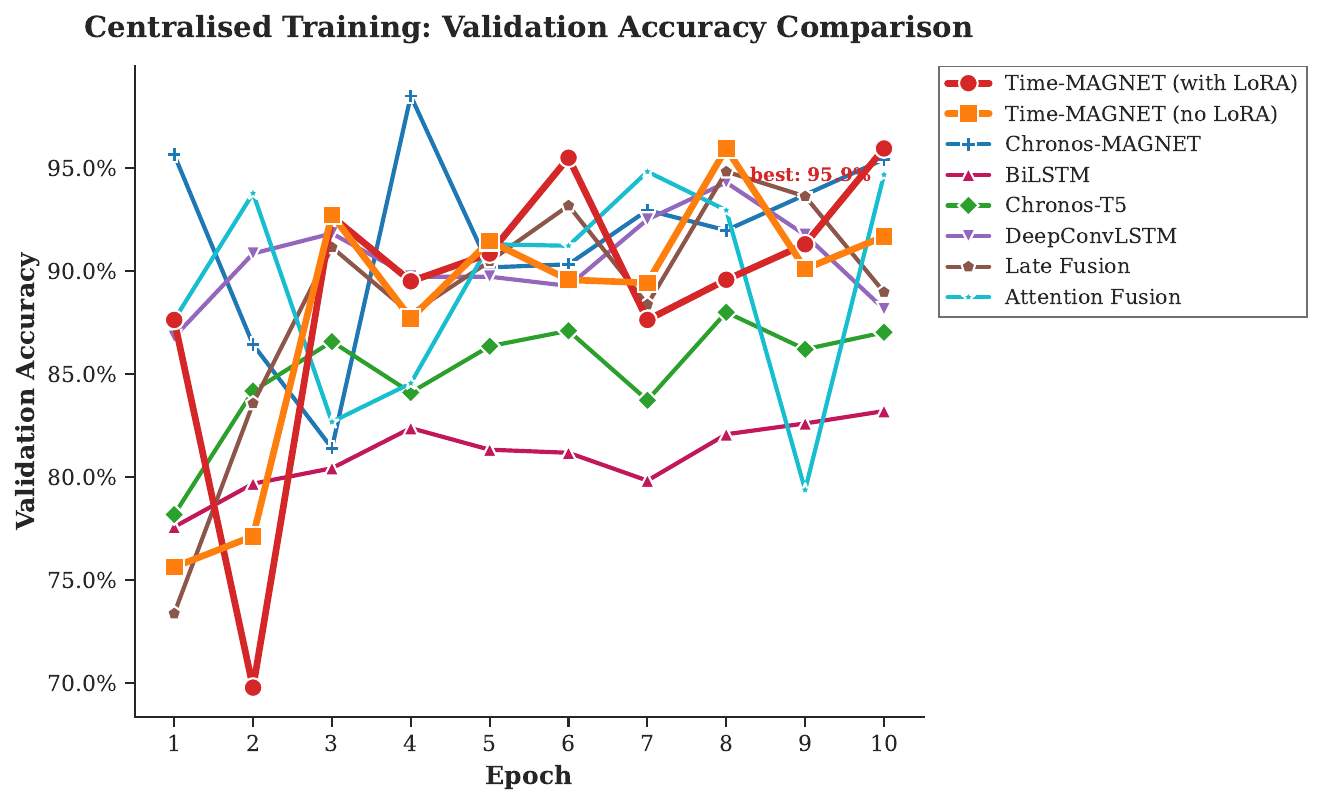}
    \caption{Validation accuracy under centralised training, comparing Time-MAGNET (with and without LoRA) against five baseline architectures over 10 epochs.}
    \label{fig3}
  \end{subfigure}
  \begin{subfigure}{0.45\textwidth}
    \centering
    \includegraphics[width=\linewidth]{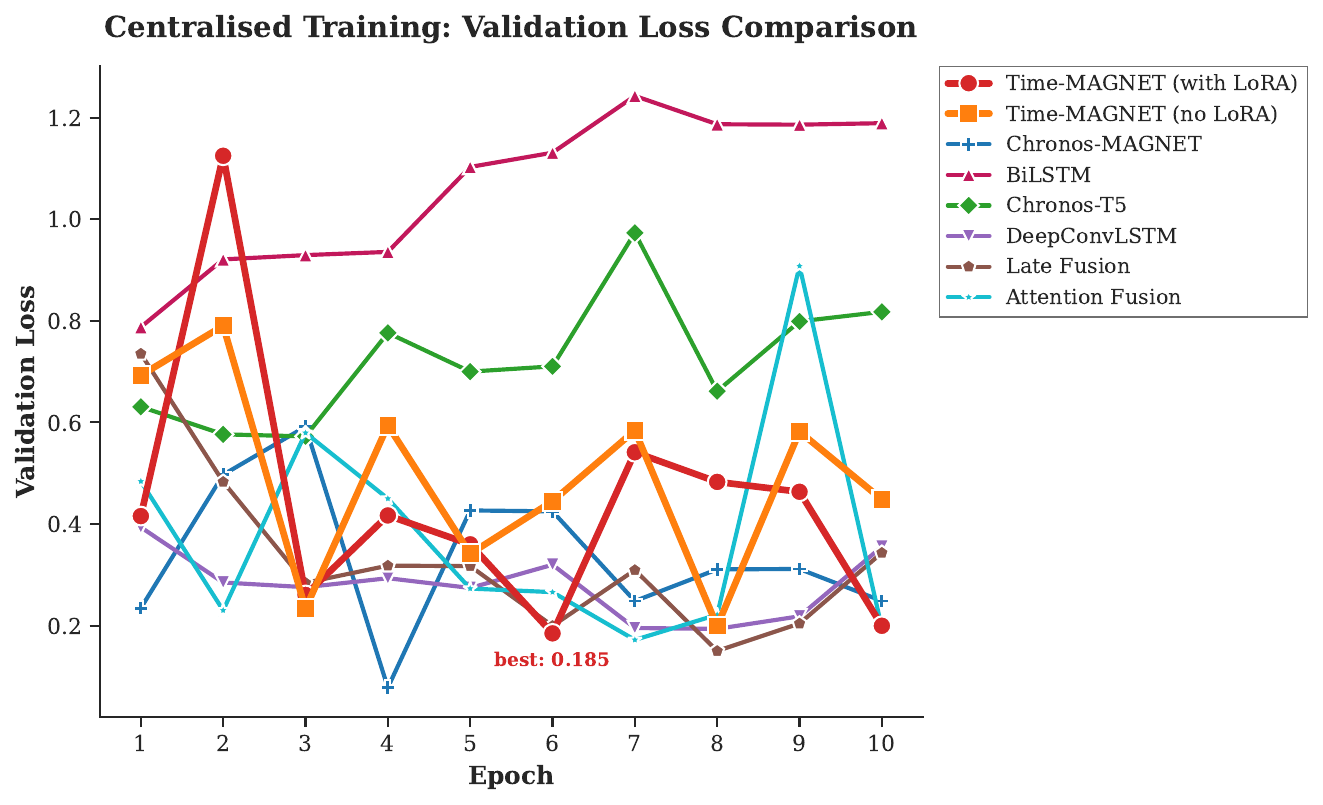}
    \caption{Validation loss under centralised training, comparing Time-MAGNET (with and without LoRA) against five baseline architectures over 10 epochs.}
    \label{fig4}
  \end{subfigure}
  % Row 2 — Federated training (FedPAIN): Time-MAGNET vs. baseline architectures
  \begin{subfigure}{0.45\textwidth}
    \centering
    \includegraphics[width=\linewidth]{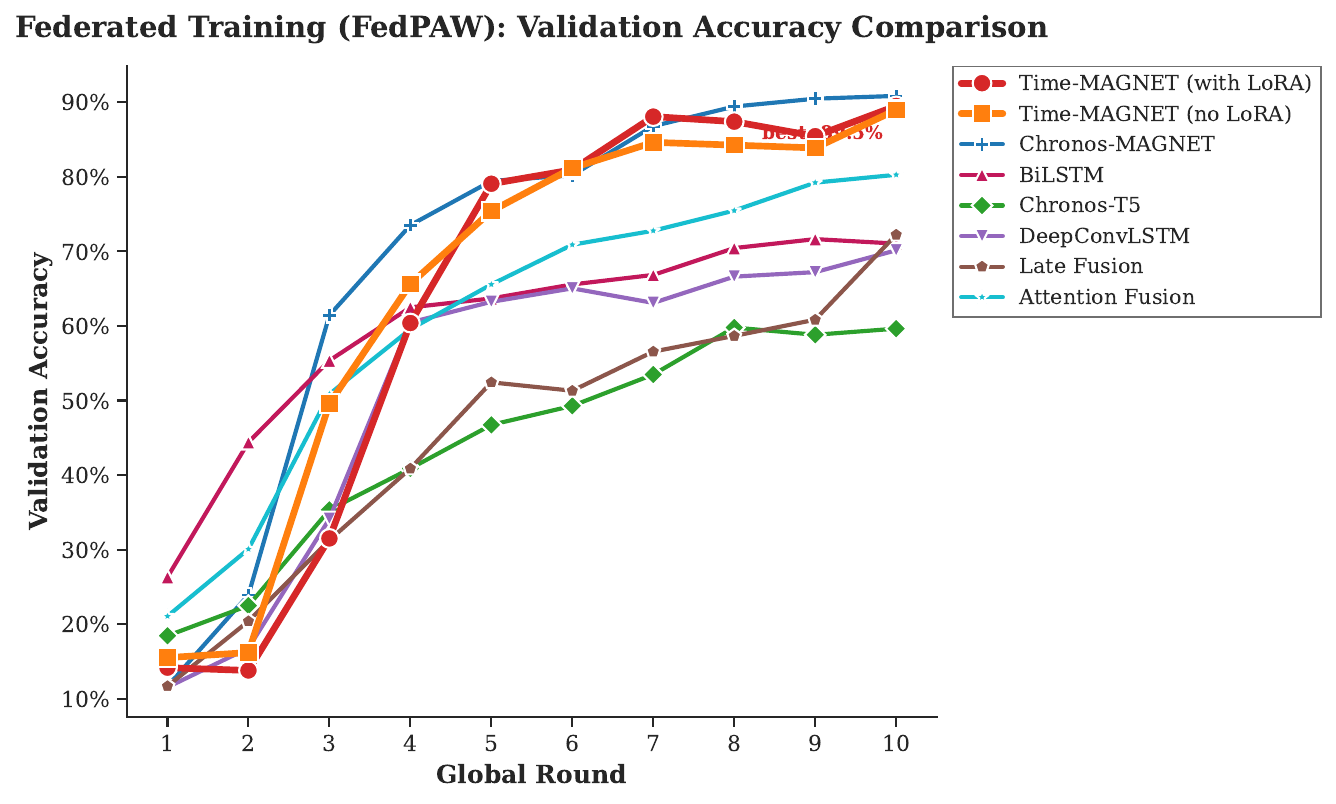}
    \caption{Validation accuracy under federated training with our proposed FedPAW algorithm, comparing Time-MAGNET (with and without LoRA) against five baseline architectures over 10 global rounds.}
    \label{fig5}
  \end{subfigure}
  \begin{subfigure}{0.45\textwidth}
    \centering
    \includegraphics[width=\linewidth]{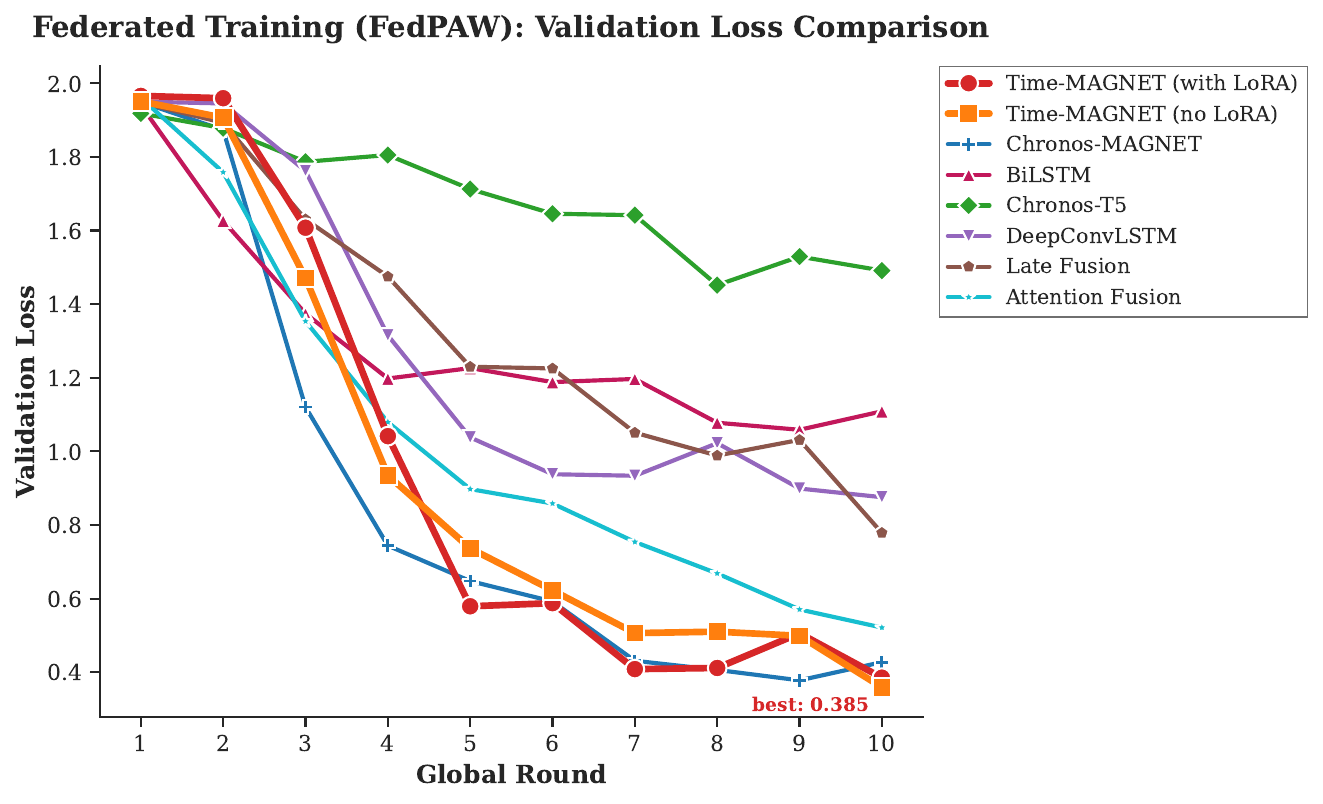}
    \caption{Validation loss under federated training with our proposed FedPAW algorithm, comparing Time-MAGNET (with and without LoRA) against five baseline architectures over 10 global rounds.}
    \label{fig6}
  \end{subfigure}
  % Row 3 — Federated algorithm comparison: Time-MAGNET only, across FL strategies
  \begin{subfigure}{0.45\textwidth}
    \centering
    \includegraphics[width=\linewidth]{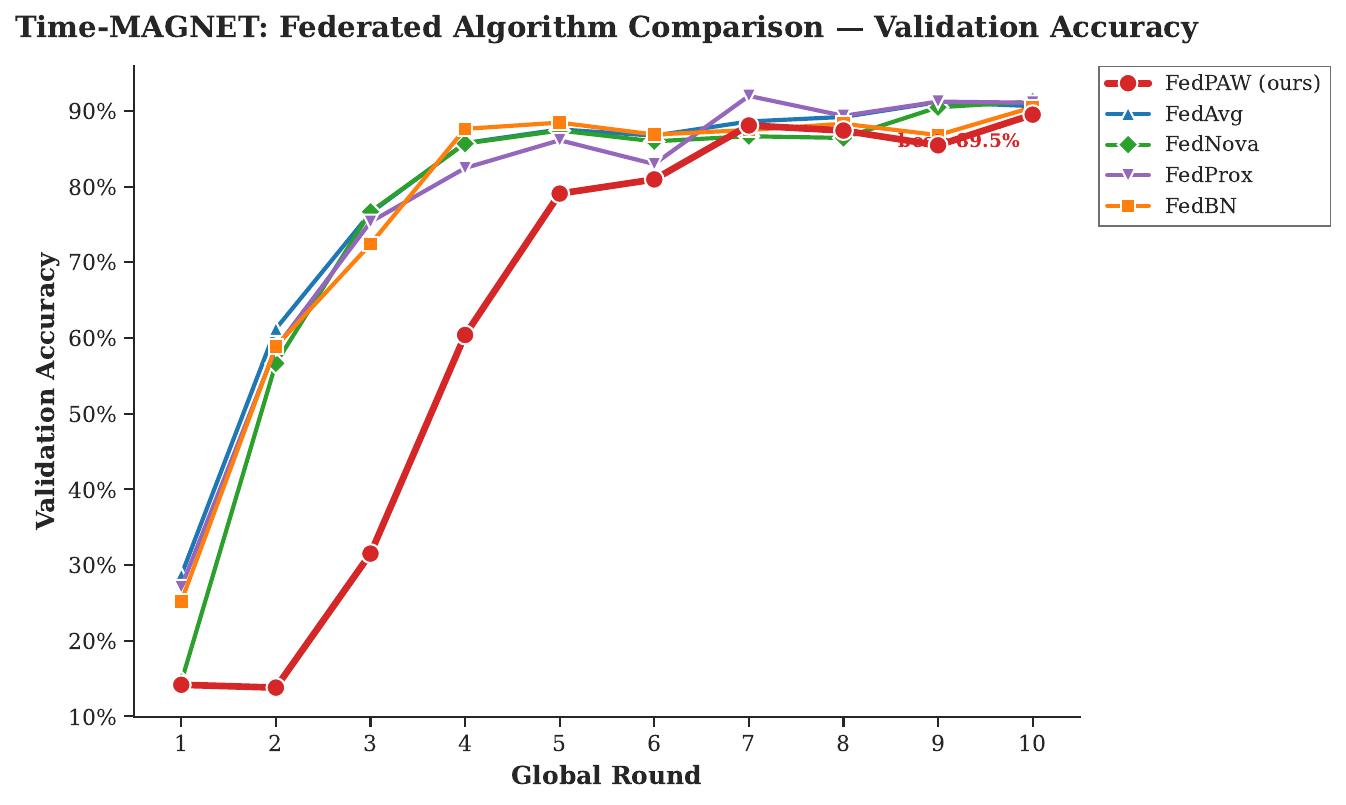}
    \caption{Validation accuracy of Time-MAGNET under our proposed FedPAW algorithm compared with four federated learning baselines (FedAvg, FedNova, FedProx, FedBN) over 10 global rounds.}
    \label{fig7}
  \end{subfigure}
  \begin{subfigure}{0.45\textwidth}
    \centering
    \includegraphics[width=\linewidth]{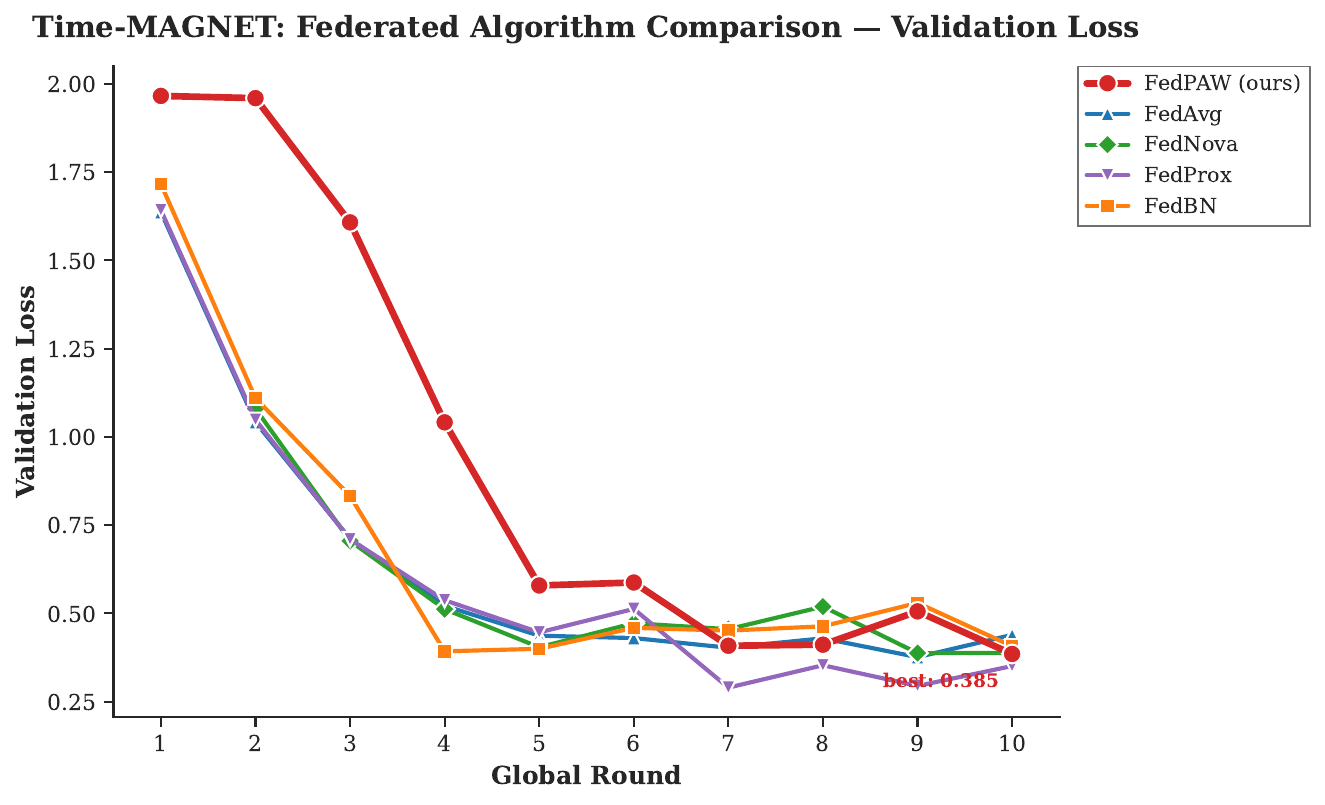}
    \caption{Validation loss of Time-MAGNET under our proposed FedPAW algorithm compared with four federated learning baselines (FedAvg, FedNova, FedProx, FedBN) over 10 global rounds.}
    \label{fig8}
  \end{subfigure}
  \caption{Validation accuracy (left column) and validation loss (right column) across three experimental settings. \textbf{Top:} centralised training, comparing Time-MAGNET against five baseline architectures. \textbf{Middle:} federated training under our proposed FedPAW algorithm, comparing the same set of architectures. \textbf{Bottom:} federated algorithm comparison isolating Time-MAGNET's performance under FedPAW against four established federated learning strategies (FedAvg, FedNova, FedProx, FedBN).}
  \label{figAccLoss}
\end{figure*}

\subsection{Convergence Behaviour}

Figure~\ref{figAccLoss} presents the validation accuracy and loss across the three experimental settings. Under centralised training (Figure~\ref{figAccLoss}a,b), Time-MAGNET with LoRA achieves the lowest validation loss (0.185) and maintains accuracy above 90\% from epoch~6 onward. In contrast, BiLSTM shows an increasing validation loss over training, suggesting weaker generalisation on this task. The other models exhibit moderate fluctuations in their validation curves rather than a smooth decrease. These variations may partly reflect the limited size of the validation set.

Under federated training with FedPAW (Figure~\ref{figAccLoss}(c,d)), all models initially exhibit accuracy close to the chance level of 14.3\% expected under uniform random guessing across seven classes, with validation loss near $\ln 7 \approx 1.95$. As training progresses, Time-MAGNET with LoRA achieves the lowest final validation loss (0.385) and the highest final accuracy among the eight architectures. Its initial improvement is slower than that of several baselines, but its performance increases substantially in later rounds.
% We examine this convergence pattern further in Section~\ref{sec:fedpain-novel}.

Figures~\ref{figAccLoss}~(e) and (f) show the convergence of Time-MAGNET under the five federated optimization methods. FedAvg, FedNova, and FedBN exhibit faster initial progress, reaching approximately 55-65\% validation accuracy by round~3. FedPAW, in contrast, remains near the initial accuracy level through the first three rounds before improving sharply between rounds~4 and~7. It overtakes the competing methods around rounds~7-8 and achieves the best final accuracy and lowest validation loss. Although this delayed improvement comes with slower initial convergence, FedPAW delivers the strongest final performance among the methods considered.

% \subsection{Performance Comparison}

% Table~\ref{tab:all_models} reports accuracy and F1 (mean $\pm$ standard deviation over repeated runs) for all eight architectures in both training regimes. Time-MAGNET with LoRA achieves the best result overall, with a centralised F1 of 0.9550 $\pm$ 0.0020 and a federated F1 of 0.8983 $\pm$ 0.0014. The gap between centralised and federated performance for this model (6.0 percentage points of F1) is the smallest such gap among all eight architectures, indicating that Time-MAGNET degrades the least when moving from the centralised to the federated setting.

\begin{table*}[h!]
\caption{Performance Comparison of Centralized and Federated  Models (values reported as mean $\pm$ standard deviation)}
\label{tab:all_models}
\centering
\begin{tabular}{lccccc}
    \toprule
    \multirow{2}{*}{Model} & \multicolumn{2}{c}{Centralized} & \multicolumn{2}{c}{Federated} \\
    \cmidrule(lr){2-3} \cmidrule(lr){4-5}
    & Accuracy & F1 & Accuracy & F1 \\
    \midrule
    Time-MAGNET(with LoRA) & 0.9555 $\pm$ 0.0018 & 0.9550 $\pm$ 0.0020 & 0.8997 $\pm$ 0.0017 & 0.8983 $\pm$ 0.0014 \\
    \midrule
    Time-MAGNET(no LoRA) & 0.9217 $\pm$ 0.0013 & 0.9203 $\pm$ 0.0010 & 0.8834 $\pm$ 0.0020 & 0.8830 $\pm$ 0.0016 \\
    \midrule
    Chronos-T5 & 0.8256 $\pm$ 0.0026 & 0.8168 $\pm$ 0.0032 & 0.6028 $\pm$ 0.0021 & 0.5786 $\pm$ 0.0023 \\
    \midrule
    Chronos-MAGNET & 0.9346 $\pm$ 0.0026 & 0.9338 $\pm$ 0.0032 & 0.8680 $\pm$ 0.0021 & 0.8660 $\pm$ 0.0023 \\
    \midrule
    BiLSTM  & 0.7939 $\pm$ 0.0021 & 0.7902 $\pm$ 0.0017 & 0.7107 $\pm$ 0.0023 & 0.6941 $\pm$ 0.0018 \\
    \midrule
    DeepConvLSTM & 0.8722 $\pm$ 0.0011 & 0.8718 $\pm$ 0.0015 & 0.6878 $\pm$ 0.0019 & 0.6724 $\pm$ 0.0015 \\
    \midrule
    Late Fusion & 0.9101 $\pm$ 0.0014 & 0.9081 $\pm$ 0.0017 & 0.7852 $\pm$ 0.0033 & 0.7880 $\pm$ 0.0028 \\
    \midrule
    Attention Fusion  & 0.9221 $\pm$ 0.0018 & 0.9220 $\pm$ 0.0016 & 0.8622 $\pm$ 0.0026 & 0.8610 $\pm$ 0.0023 \\
    \bottomrule
\end{tabular}
\end{table*}

\subsection{Performance Comparison}

Table~\ref{tab:all_models} reports the mean accuracy and F1-score, along with their standard deviations, for all eight architectures under centralised and federated training. Time-MAGNET with LoRA achieves the highest performance in both settings, with an accuracy of $0.9555$ and an F1-score of $0.9550$ under centralised training, and an accuracy of $0.8997$ and an F1-score of $0.8983$ under federated training. Although its F1-score decreases by 5.67 percentage points in the federated setting, it retains the highest federated performance among all eight architectures.

Chronos-MAGNET is the closest competitor, achieving F1-scores of 0.9338 and 0.8660 under centralised and federated training, respectively. Thus, Time-MAGNET with LoRA improves upon it by 2.12 percentage points centrally and 3.23 percentage points under federated training. Since both models use the same MAGNET fusion mechanism but different T5 encoders, this comparison highlights the importance of the encoder design and adaptation strategy for this task. We examine the effect of LoRA in more detail in Section~\ref{sec:lora-novel}.

The performance differences are also substantial when compared with Chronos-T5, BiLSTM, and DeepConvLSTM. Relative to Time-MAGNET with LoRA, Chronos-T5 has lower F1-scores by 13.82 percentage points under centralised training and 31.97 percentage points under federated training. The corresponding differences are 16.48 and 20.42 percentage points for BiLSTM, and 8.32 and 22.59 percentage points for DeepConvLSTM. These results indicate that the proposed architecture performs better than both the individual time-series encoder and the conventional recurrent and convolutional-recurrent baselines, with particularly pronounced gains under federated training.

\paragraph{\textbf{Effect of the Fusion Mechanism:}} Late Fusion and Attention Fusion achieve intermediate performance. Their centralised F1-scores are 0.9081 and 0.9220, respectively, while their federated F1-scores are 0.7880 and 0.8610. Time-MAGNET with LoRA outperforms both methods in the two training settings. Since all three models use the same T5 and DART-CNN encoders, this comparison highlights the effect of the fusion strategy. Late Fusion combines the modality-specific features through concatenation, whereas Attention Fusion uses conventional multi-head self-attention. In contrast, MAGNET explicitly models inter-modal relationships through graph attention, alongside mixture-of-experts processing and skip connections. The higher F1-scores of Time-MAGNET suggest that this structured fusion approach is more effective for multimodal HAR, particularly under federated training.

The results for BiLSTM and DeepConvLSTM provide further context. Although both models capture temporal dependencies, they do not explicitly model inter-modal relationships through a graph-based fusion mechanism. DeepConvLSTM also incurs the highest training time among the eight models (Table~\ref{tab:model_complexity}). These results suggest that the performance advantage of Time-MAGNET is related not only to temporal representation learning but also to its structured approach to multimodal fusion.

\subsection{Effect of LoRA}
\label{sec:lora-novel}

In both variants, the T5 encoder is randomly initialized without using pretrained weights. In Time-MAGNET with LoRA, the encoder's original parameters are kept frozen, and trainable rank-16 LoRA adapters are introduced into its query, key, and value projections. Thus, the encoder learns through updates to these adapters rather than through direct updates to its original weights. In Time-MAGNET without LoRA, \textit{all T5 encoder parameters are updated directly during training}. In both variants, the remaining components, including DART-CNN, MAGNET, and the classification head, are trained normally.

As shown in Table~\ref{tab:all_models}, LoRA improves the F1-score from 0.9203 to 0.9550 under centralised training and from 0.8830 to 0.8983 under federated training. It also reduces the number of trainable parameters from 55.83M to 47.54M (a 14.85\% reduction). As reported in Table~\ref{tab:model_complexity}, training time decreases from 3714.76 to 3485.05 seconds under centralised training and from 8729.16 to 8052.36 seconds under federated training.

% As shown in Table~\ref{tab:all_models}, LoRA improves the F1-score from 0.9203 to 0.9550 under centralised training and from 0.8830 to 0.8983 under federated training. It also reduces the number of trainable parameters from approximately 55.83M to 47.54M, a reduction of 14.85\%. Training time decreases from 3714.76 to 3485.05 seconds in the centralised setting and from 8729.16 to 8052.36 seconds in the federated setting see Table~\ref{tab:model_complexity}.

These results suggest that adapting the randomly initialized T5 encoder through low-rank attention updates can be more effective than full encoder fine-tuning for this task. One possible explanation is that restricting updates to the attention projections provides a useful regularisation effect, while allowing the remaining network components to learn the multimodal representations. The improvement in both training regimes indicates that LoRA can be useful beyond its conventional role in adapting pretrained models.

\begin{table*}[h!]
\caption{Performance comparison of Time-MAGNET model under different federated optimization algorithms (values reported as mean $\pm$ standard deviation).}
    \label{tab_time_mag_fed}
    \centering
    \begin{tabular}{lccc}
        \toprule
        Algorithm & Accuracy & F1 & Training Time (s)\\
        \midrule
        FedAvg   & 0.8693 $\pm$ 0.0012 & 0.8689 $\pm$ 0.0015 & 6967.53 \\
        FedProx  & 0.8847 $\pm$ 0.0010 & 0.8843 $\pm$ 0.0009 & 8023.40 \\
        FedNova  & 0.8672 $\pm$ 0.0018 & 0.8661 $\pm$ 0.0021 & 6913.21\\
        FedBN    & 0.8693 $\pm$ 0.0027 & 0.8689 $\pm$ 0.0023 & 6974.14\\
        FedPAW  & 0.8997 $\pm$ 0.0015 & 0.8983 $\pm$ 0.0013 & 8052.36 \\
        \bottomrule
    \end{tabular}
\end{table*}

\subsection{Federated Optimizer Comparison}
\label{sec:fedpain-novel}

Table~\ref{tab_time_mag_fed} compares five federated optimization algorithms while keeping the Time-MAGNET architecture fixed. FedPAW achieves the best performance, with an accuracy of 0.8997 and an F1-score of 0.8983. Compared with the next-best method, FedProx, it improves accuracy by 1.50 percentage points and F1-score by 1.40 percentage points. The gains over FedAvg, FedNova, and FedBN are larger, at approximately 3.0 percentage points in accuracy and 2.9 percentage points in F1-score.

These improvements can be attributed to FedPAW's combination of adaptive client weighting and damped server updates. Its client weights account for both the effective number of local optimization steps and the logarithm of the local dataset size, reducing the influence of clients solely because they have larger datasets. The server then aggregates the weighted client updates and applies damping before updating the global model. Unlike FedProx, which controls local client drift through a proximal term, FedPAW incorporates this additional server-side damping. This combination is designed to moderate the effect of heterogeneous client updates across communication rounds.

The convergence curves in Figure~\ref{figAccLoss} \,(e) and (f) show that FedPAW improves more slowly during the initial rounds but achieves the highest final performance, overtaking the competing methods around rounds 7--8. This behaviour is consistent with the conservative nature of its damped server updates. In terms of computational cost, FedPAW requires 8052.36 seconds, which is close to FedProx (8023.40 seconds) but approximately 15--16\% higher than FedAvg, FedNova, and FedBN. Thus, FedPAW achieves better predictive performance at a modest additional training cost.

\begin{figure*}[htbp]
  \centering
  % Top row
  \begin{subfigure}{0.45\textwidth}
    \centering
    \includegraphics[width=\linewidth]{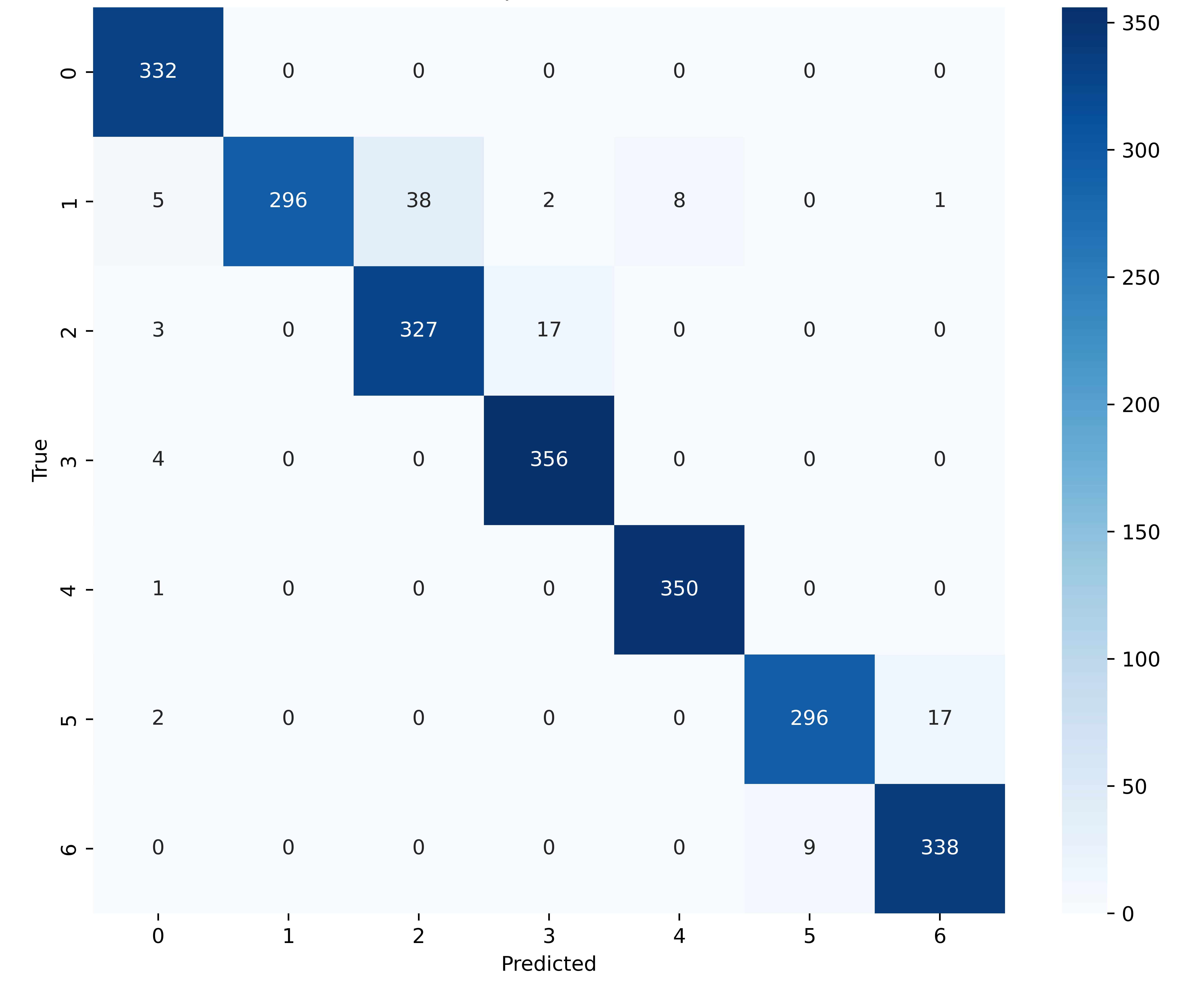}
    \caption{Centralised Time-MAGNET with LoRA}
    \label{fig9}
  \end{subfigure}
  \begin{subfigure}{0.45\textwidth}
    \centering
    \includegraphics[width=\linewidth]{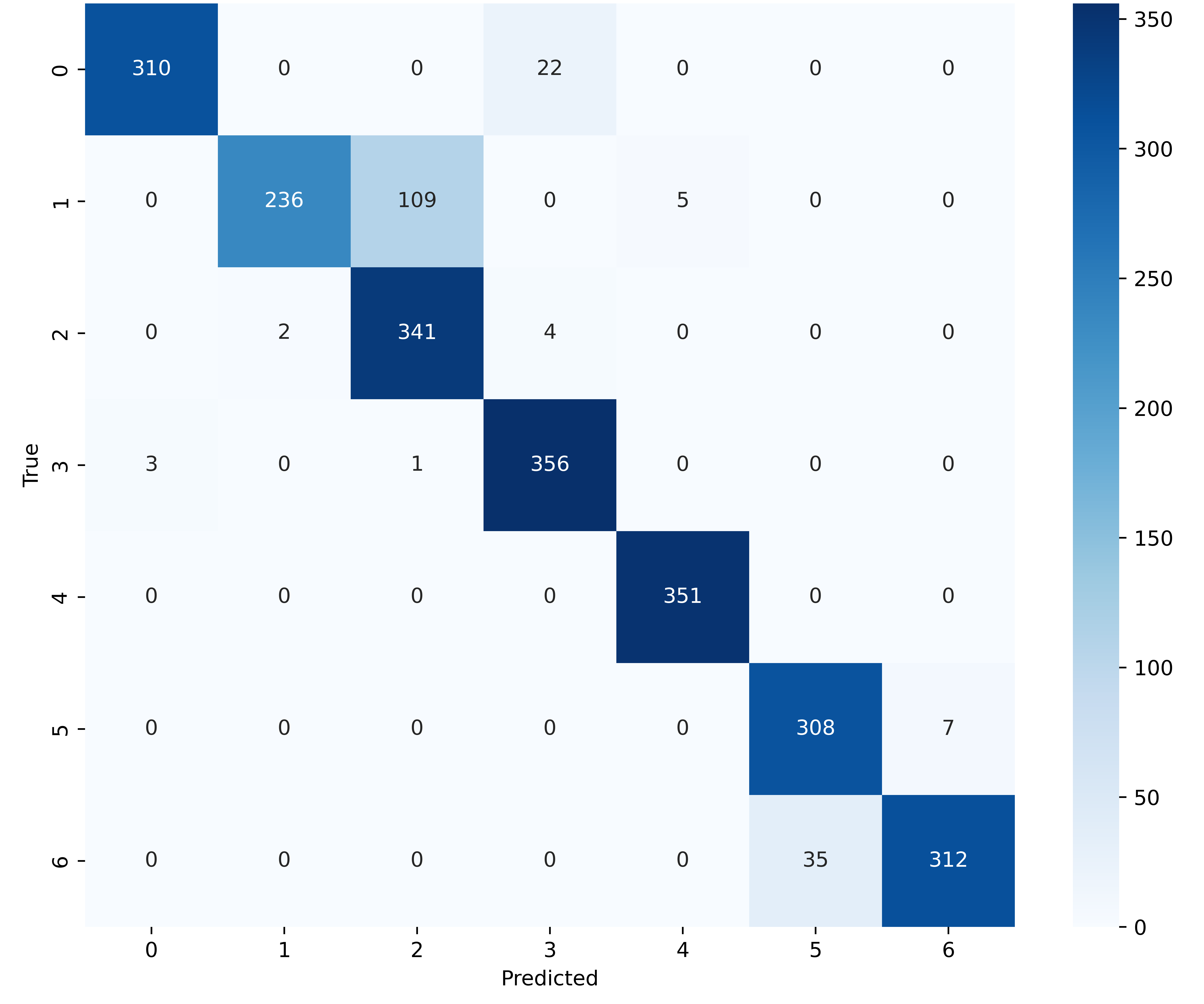}
    \caption{Centralised Time-MAGNET without LoRA}
    \label{fig10}
  \end{subfigure}
  % Bottom row
  \begin{subfigure}{0.45\textwidth}
    \centering
    \includegraphics[width=\linewidth]{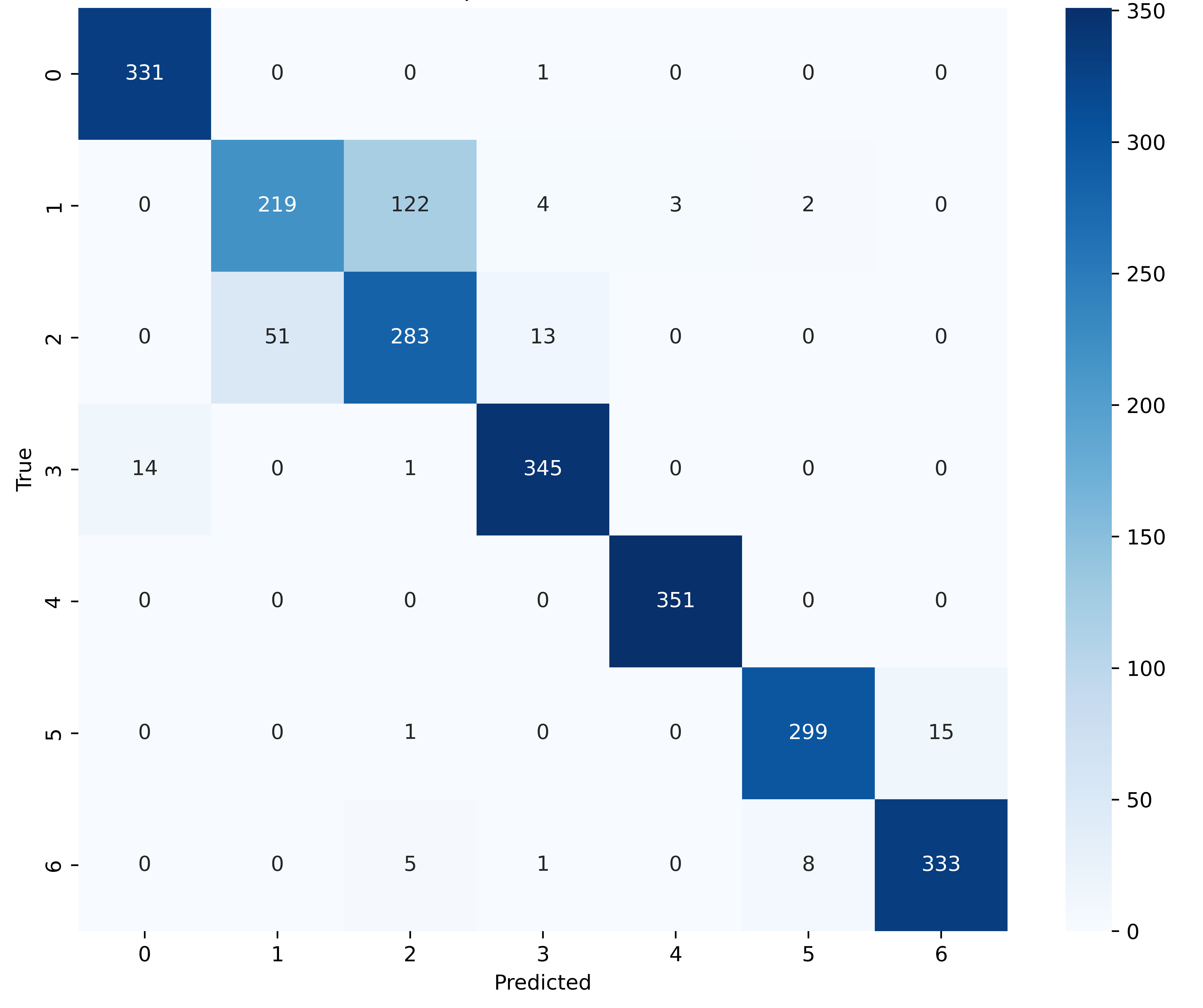}
    \caption{Federated (FedPAW) Time-MAGNET with LoRA}
    \label{fi11}
  \end{subfigure}
  \begin{subfigure}{0.45\textwidth}
    \centering
    \includegraphics[width=\linewidth]{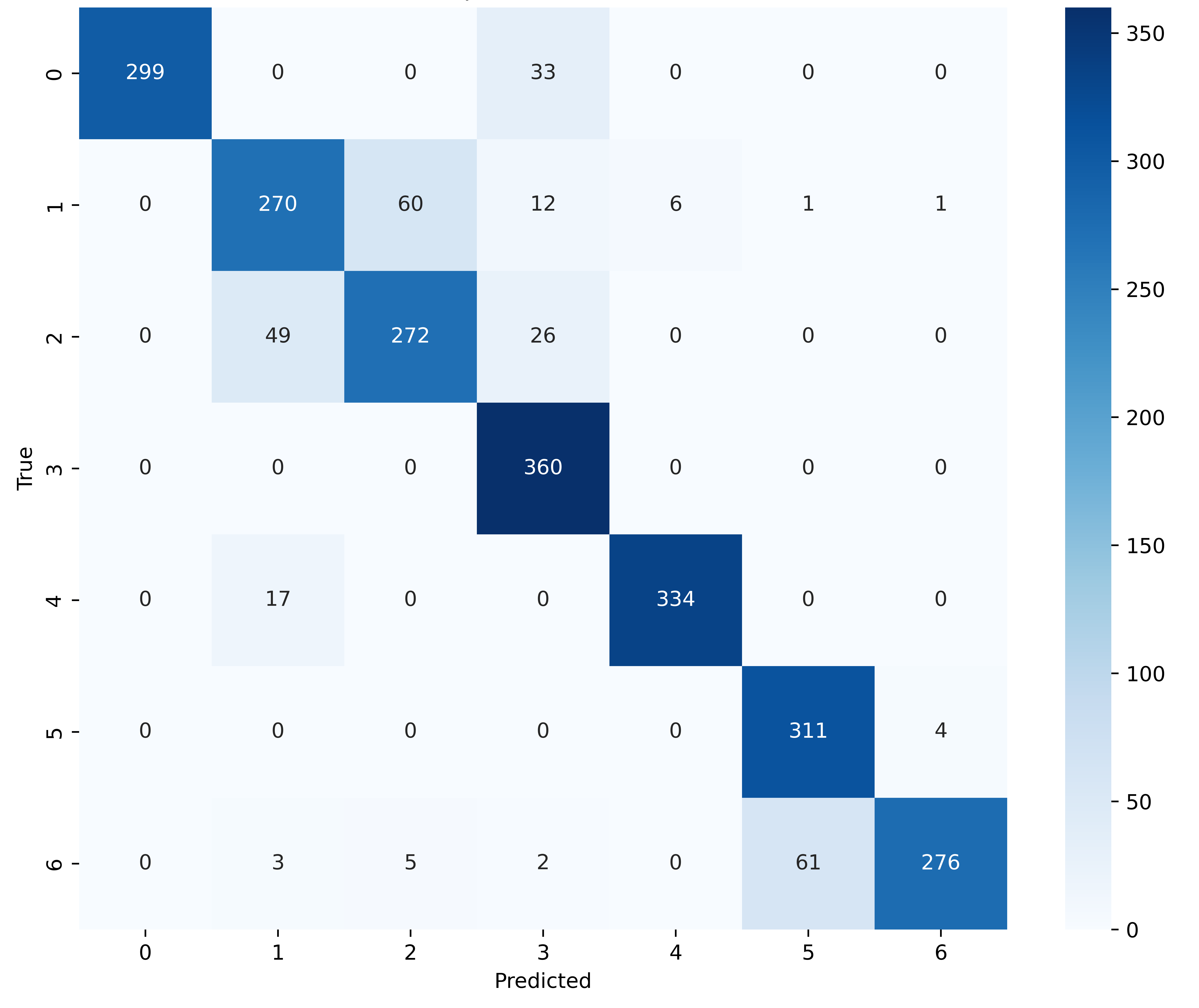}
    \caption{Federated (FedPAW) Time-MAGNET without LoRA}
    \label{fig12}
  \end{subfigure}
  \caption{Confusion matrices for Time-MAGNET across training paradigms and LoRA configurations. \textbf{Top:} centralised training, with (left) and without (right) LoRA. \textbf{Bottom:} federated training under our proposed FedPAW algorithm, with (left) and without (right) LoRA. Each matrix reports per-class prediction counts on the held-out validation set, allowing class-wise error patterns to be compared across settings.}
  \label{figCM}
\end{figure*}

\begin{figure*}[htbp]
  \centering
  % Top row
  \begin{subfigure}{0.45\textwidth}
    \centering
    \includegraphics[width=\linewidth]{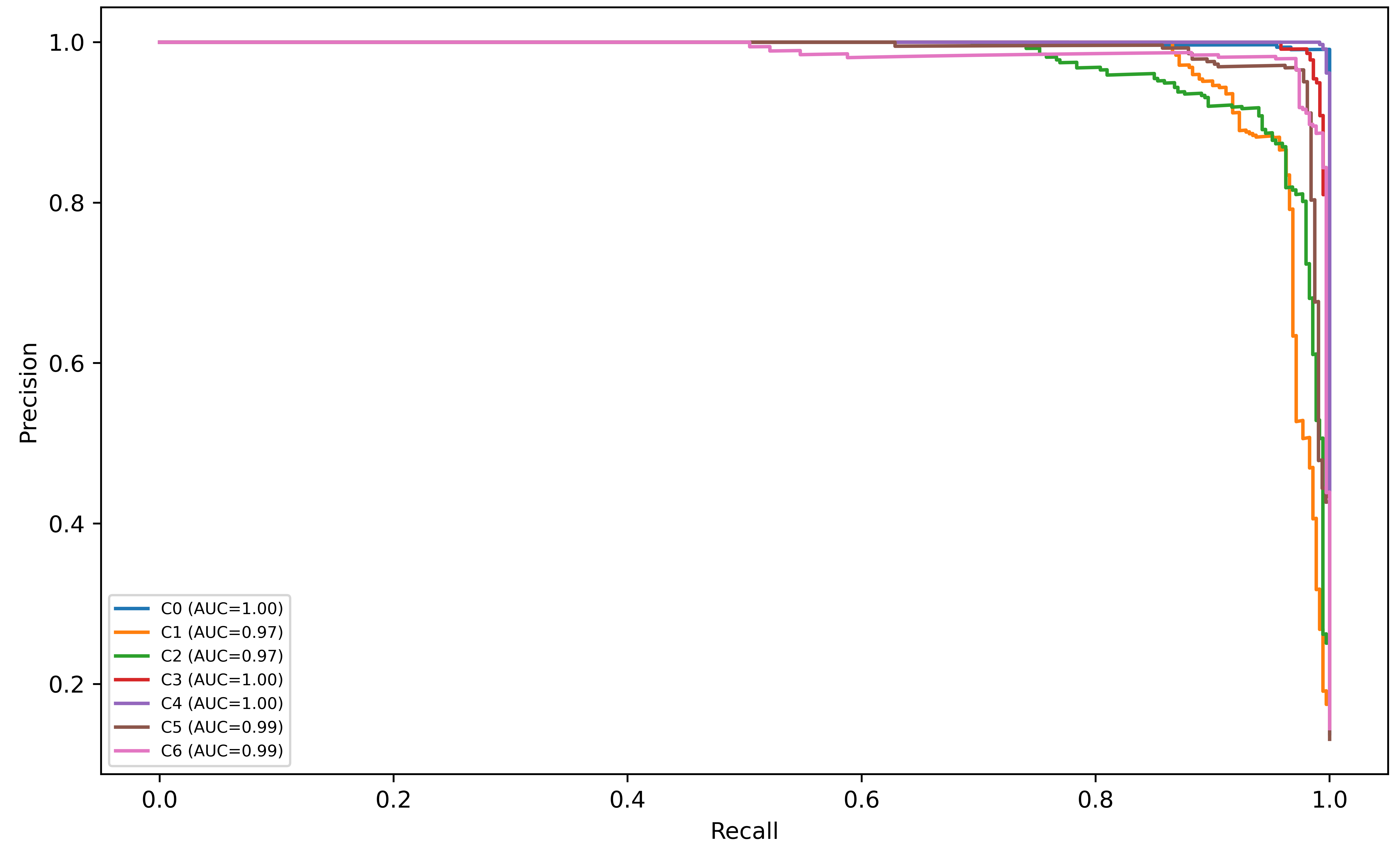}
    \caption{Centralised Time-MAGNET with LoRA}
    \label{fig13}
  \end{subfigure}
  \begin{subfigure}{0.45\textwidth}
    \centering
    \includegraphics[width=\linewidth]{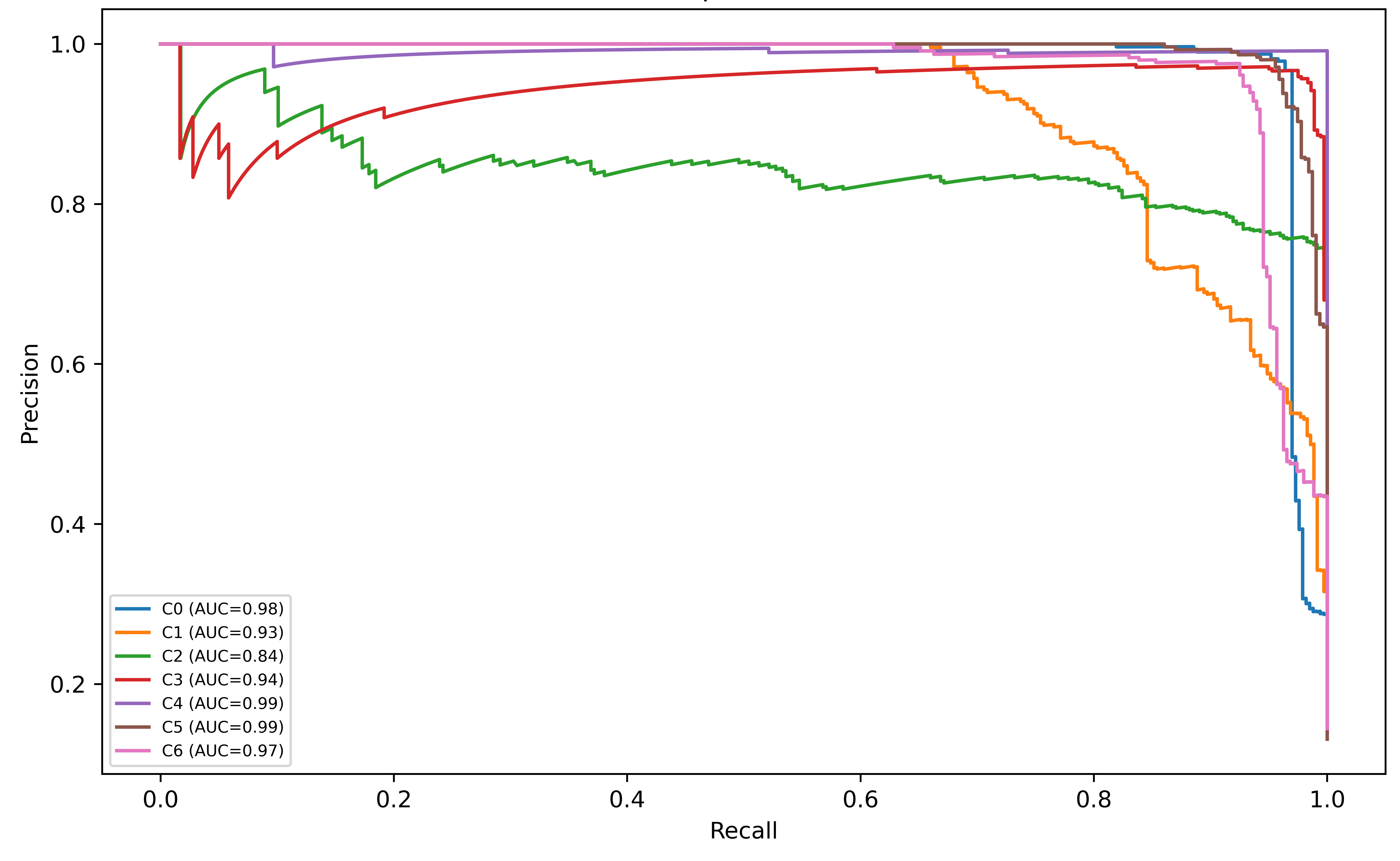}
    \caption{Centralised Time-MAGNET without LoRA}
    \label{fig14}
  \end{subfigure}
  % Bottom row
  \begin{subfigure}{0.45\textwidth}
    \centering
    \includegraphics[width=\linewidth]{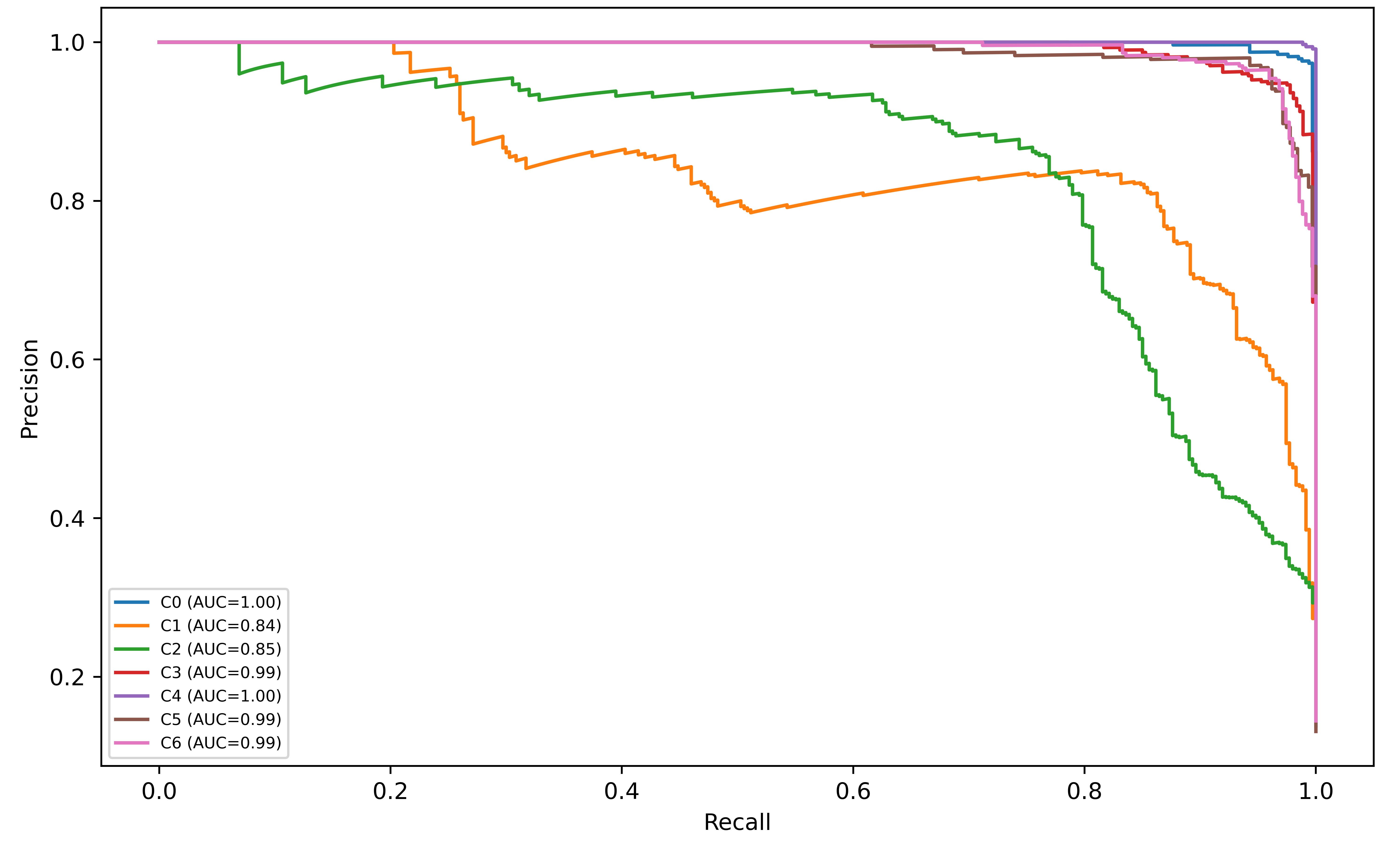}
    \caption{Federated (FedPAW) Time-MAGNET with LoRA}
    \label{fig15}
  \end{subfigure}
  \begin{subfigure}{0.45\textwidth}
    \centering
    \includegraphics[width=\linewidth]{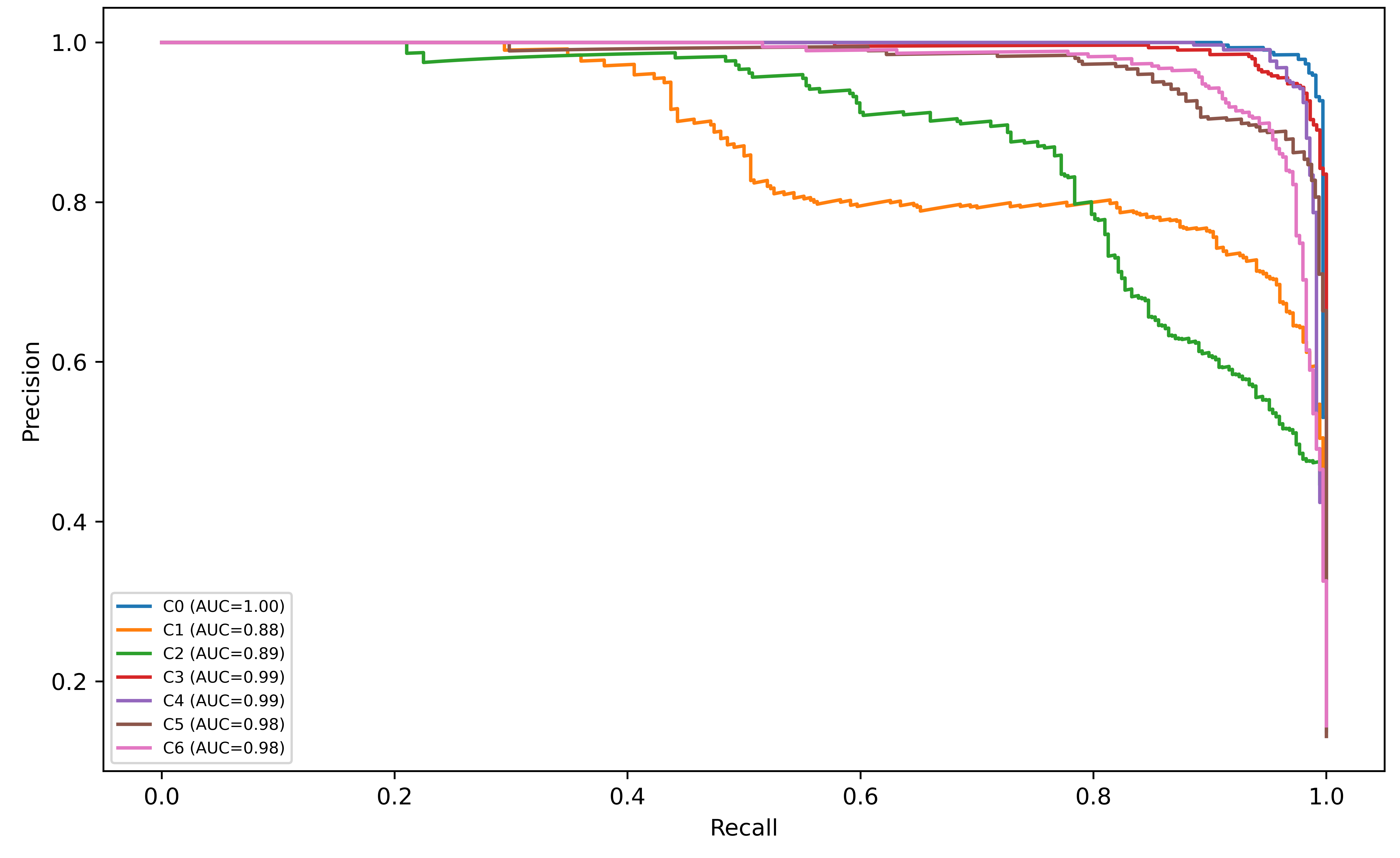}
    \caption{Federated (FedPAW) Time-MAGNET without LoRA}
    \label{fig16}
  \end{subfigure}
  \caption{Precision-recall curves for Time-MAGNET across training paradigms and LoRA configurations. \textbf{Top:} centralised training, with (left) and without (right) LoRA. \textbf{Bottom:} federated training under our proposed FedPAW algorithm, with (left) and without (right) LoRA. These curves characterise the precision-recall trade-off at each classification threshold, complementing the accuracy and loss curves reported earlier.}
  \label{figPR}
\end{figure*}

\begin{figure*}[htbp]
  \centering
  % Top row
  \begin{subfigure}{0.45\textwidth}
    \centering
    \includegraphics[width=\linewidth]{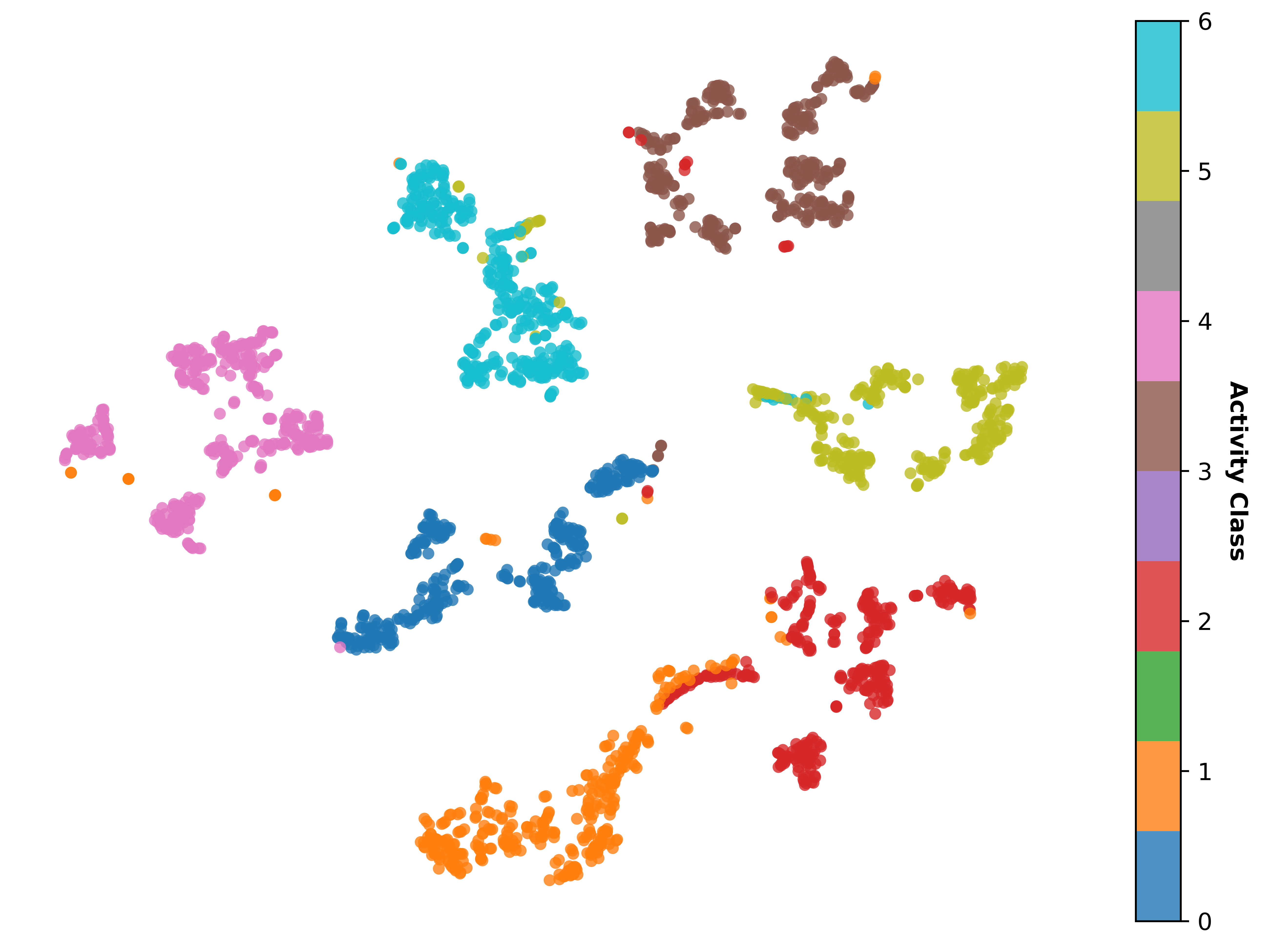}
    \caption{Centralised Time-MAGNET with LoRA}
    \label{fig17}
  \end{subfigure}
  \begin{subfigure}{0.45\textwidth}
    \centering
    \includegraphics[width=\linewidth]{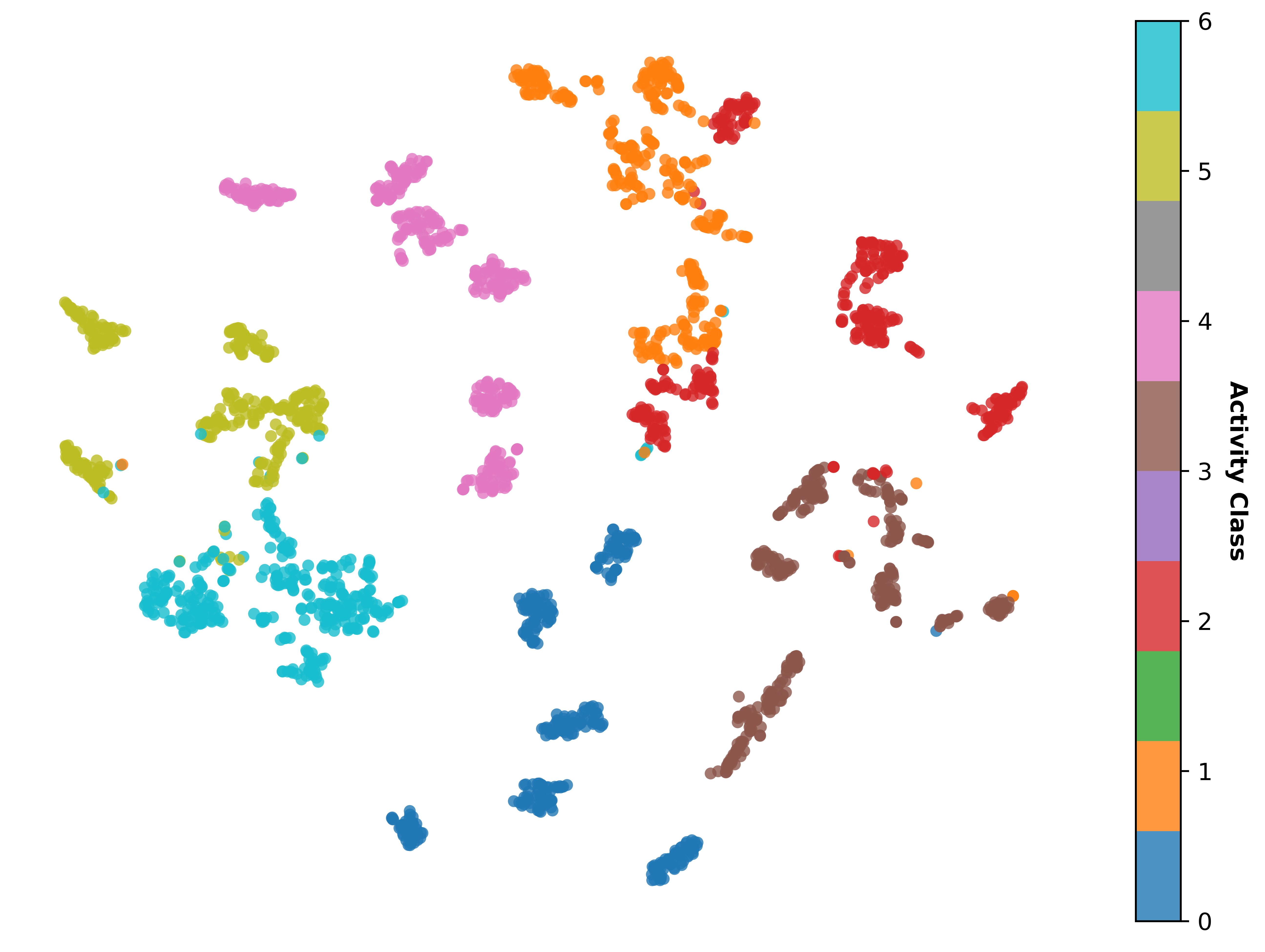}
    \caption{Federated (FedPAW) Time-MAGNET with LoRA}
    \label{fig18}
  \end{subfigure}
  % Bottom row
  \begin{subfigure}{0.45\textwidth}
    \centering
    \includegraphics[width=\linewidth]{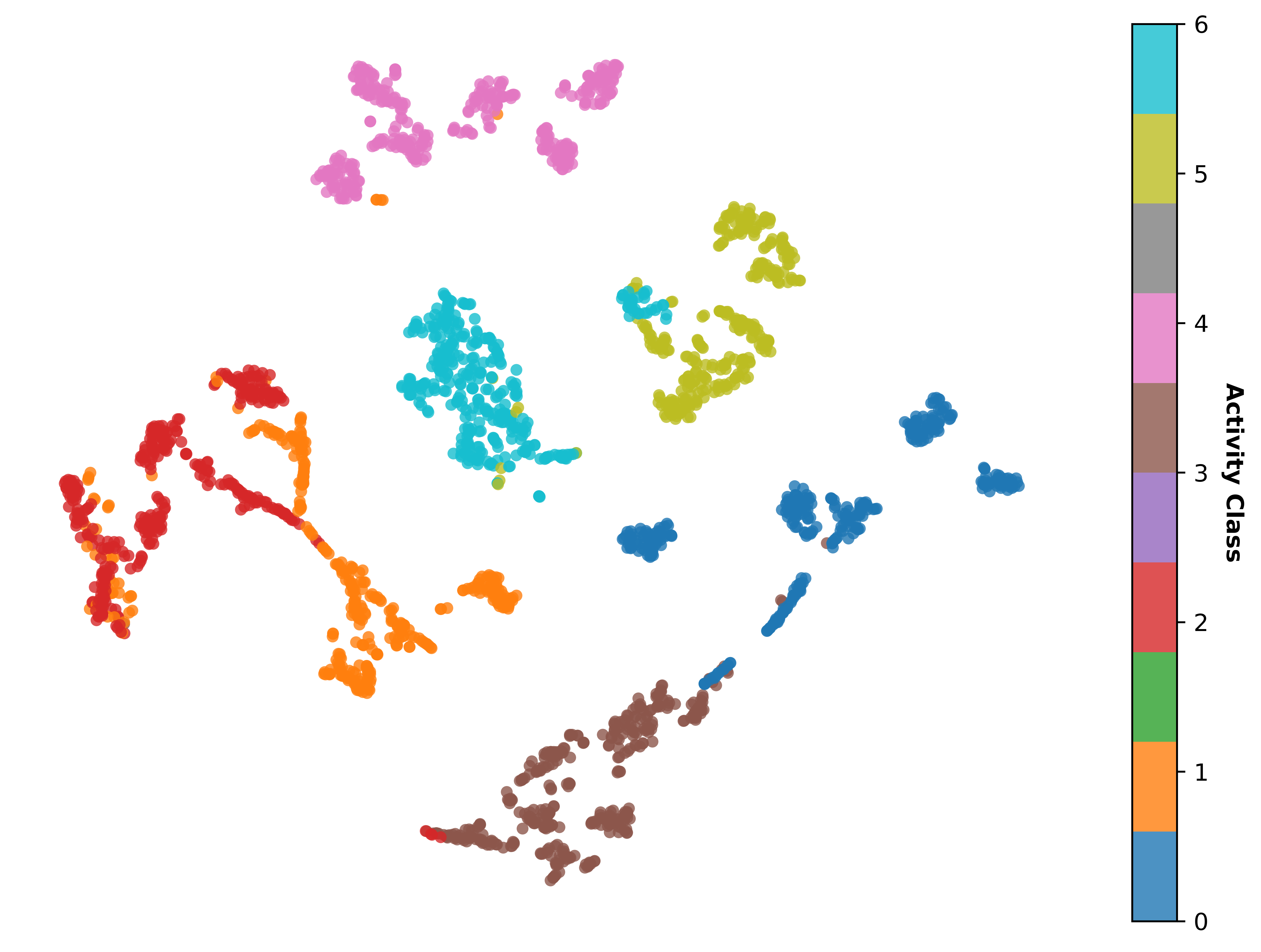}
    \caption{Centralised Time-MAGNET without LoRA}
    \label{fig19}
  \end{subfigure}
  \begin{subfigure}{0.45\textwidth}
    \centering
    \includegraphics[width=\linewidth]{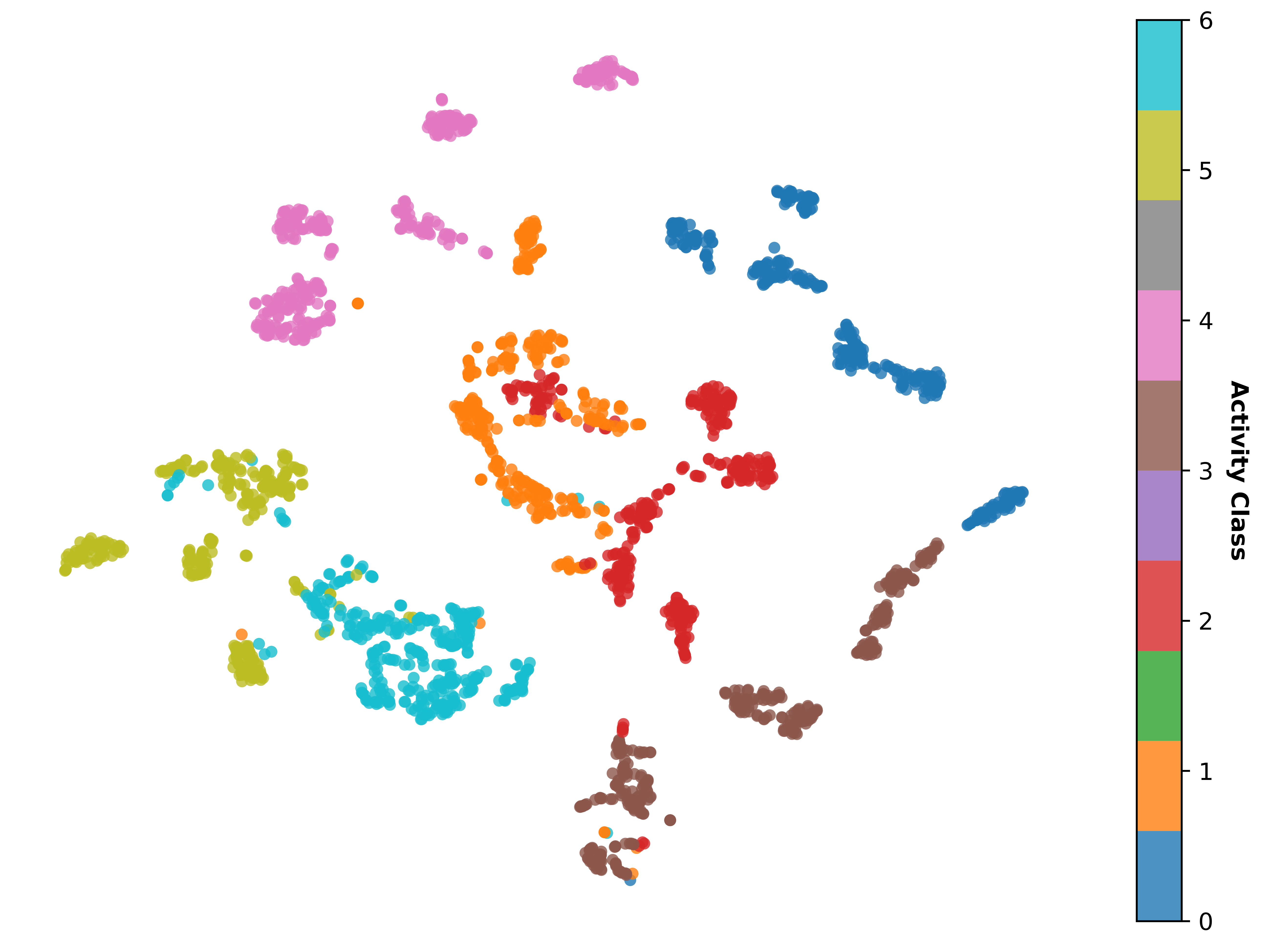}
    \caption{Federated (FedPAW) Time-MAGNET without LoRA}
    \label{fig20}
  \end{subfigure}
  \caption{t-SNE projections of the fused feature embeddings learned by Time-MAGNET. \textbf{Top:} models trained with LoRA, centralised (left) and federated under FedPAW (right). \textbf{Bottom:} models trained without LoRA, centralised (left) and federated under FedPAW (right). Rows are grouped by LoRA configuration rather than by training paradigm, letting the centralised-vs-federated embedding structure be compared directly within each row while holding LoRA use fixed.}
  \label{figTSNE}
\end{figure*}

% The t-SNE projections of the fused embeddings (Figure~\ref{figTSNE}) provide a qualitative, representation level view consistent with the quantitative findings above. The centralised, LoRA enabled embedding (Figure~\ref{figTSNE}a) forms the most compact and well separated clusters of the four configurations. Consistent with the confusion matrix and PR-AUC results, the Bridging and Pelvic tilt clusters appear adjacent with a visibly interleaved boundary, most pronounced in the centralised without LoRA embedding (Figure~\ref{figTSNE}c), a direct visual counterpart to that configuration's depressed Pelvic-tilt PR-AUC. The federated embeddings (Figure~\ref{figTSNE}b,d) show generally similar cluster topology to their centralised counterparts but with more elongated, less compact clusters and greater boundary overlap, particularly in the federated without LoRA case (Figure~\ref{figTSNE}d), which is the lowest F1 configuration among the four and correspondingly shows the most cluster overlap of the four panels. Taken together, Figures~\ref{figCM} through~\ref{figTSNE} tell a consistent story from three independent angles, raw prediction counts, threshold independent ranking quality, and the geometry of the learned representation itself, that Time-MAGNET's residual errors are concentrated in a small, physiologically explicable subset of exercise pairs rather than reflecting a general inability to learn discriminative multimodal representations.

\subsection{Class-wise Performance and Embedding Analysis}
\label{sec:classification-quality}

\paragraph{Confusion Matrices:} The dataset comprises seven physical exercises: Knee-rolling (class 0), Bridging (class 1), Pelvic tilt (class 2), The Clam (class 3), Repeated Extension in Lying (class 4), Prone punches (class 5), and Superman (class 6). We examine the confusion matrices, class-wise precision-recall (PR) curves, and fused embedding visualisations to understand the class-specific errors made by Time-MAGNET across the four configurations (centralised and federated training, each with and without LoRA).

Figure~\ref{figCM} shows that most classification errors are concentrated in a few pairs of exercises rather than being distributed uniformly across classes. The most persistent confusion occurs between Bridging (class 1) and Pelvic tilt (class 2) in all four configurations. Both exercises are performed in a supine position and involve similar pelvic and hip movements. Their distinction lies mainly in the extent and direction of pelvic movement, which may be difficult to capture from the available sensor measurements. This confusion is present even in the best performing configuration (centralised training with LoRA) and becomes more pronounced in the other configurations.

A second, less frequent confusion occurs between Prone punches (class 5) and Superman (class 6). Both exercises are performed in a prone position and involve upper-body movement, but differ in the direction and pattern of arm movement. Their similar overall body positions may contribute to the observed overlap. The confusion is most pronounced under federated training without LoRA, which also has the lowest overall F1-score among the four configurations. Another notable pattern concerns Knee-rolling (class 0) and The Clam (class 3). Confusion between these exercises appears in both configurations without LoRA but is largely absent in the LoRA enabled configurations. This observation suggests that LoRA based adaptation may help distinguish these two classes. However, the confusion matrices alone do not establish the mechanism behind this improvement. The remaining class pairs are generally better separated, indicating that the classification errors are concentrated in exercises with similar movement patterns.

\paragraph{Precision-Recall curves:} Figure~\ref{figPR} presents the class-wise precision-recall (PR) curves and their corresponding areas under the curves (PR-AUC), providing a threshold independent view of classification performance. Under centralised training with LoRA (Figure~\ref{figPR}a), all classes achieve PR-AUC values of at least 0.97. Bridging and Pelvic tilt have the lowest values (0.97), consistent with their frequent confusion in Figure~\ref{figCM}. Without LoRA (Figure~\ref{figPR}b), the PR-AUC for Pelvic tilt decreases to 0.84, while that for Bridging decreases to 0.93. This indicates a more pronounced loss in the ranking quality for Pelvic tilt in this setting.

Under federated training, Bridging and Pelvic tilt again have the lowest PR-AUC values. Their scores are 0.84 and 0.85 with LoRA (Figure~\ref{figPR}c), and 0.88 and 0.89 without LoRA (Figure~\ref{figPR}d), respectively. Thus, unlike the centralised results, removing LoRA does not further reduce the PR-AUC of these two classes under federated training. The other classes generally retain high PR-AUC values across the configurations. The PR curves support the confusion-matrix findings and show that the main classification difficulty lies in distinguishing a small number of similar exercises.

\paragraph{Embedding Analysis:} The t-SNE projections of the fused embeddings in Figure~\ref{figTSNE} provide a qualitative view of the learned class representations. The centralised configuration with LoRA (Figure~\ref{figTSNE}a) shows relatively compact and well-separated clusters. In contrast, the Bridging and Pelvic tilt clusters remain close to each other, with greater overlap in the centralised configuration without LoRA (Figure~\ref{figTSNE}c). The federated embeddings (Figures~\ref{figTSNE}b,d) also show less compact clusters and greater overlap, particularly in the configuration without LoRA. These visual patterns are broadly consistent with the confusion matrices and PR-AUC results, although t-SNE provides qualitative evidence rather than a quantitative measure of class separability.

Taken together, the three analyses show that classification errors are concentrated in exercises with similar movement patterns, with performance varying across training and LoRA configurations.

\begin{table*}[h!]
\caption{Performance Comparison of Centralized and Federated  MAGNET Models with varying modalities (values reported as mean $\pm$ standard deviation).}
    \label{tab: ablation}
    \centering
    \begin{tabular}{lcccc}
        \toprule
        \multirow{2}{*}{Modality} & \multicolumn{2}{c}{Centralized} & \multicolumn{2}{c}{Federated} \\
        \cmidrule(lr){2-3} \cmidrule(lr){4-5}
        & Accuracy & F1 & Accuracy & F1 \\
        \midrule
        {act, acw, pm} & 0.8504 $\pm$0.0016 & 0.8480 $\pm$0.0011 & 0.7306 $\pm$0.0009 & 0.7237 $\pm$0.0012 \\
        {act, acw, dc} & 0.8818 $\pm$0.0014 & 0.8816 $\pm$0.0010 & 0.8018 $\pm$0.0008 & 0.8037 $\pm$0.0012 \\
        {act, pm, dc} & 0.9184 $\pm$0.0003 & 0.9189 $\pm$0.0005 & 0.8938 $\pm$0.0007 & 0.8936 $\pm$0.0002 \\
        {acw, pm, dc} &  0.9284 $\pm$0.0006 & 0.9281 $\pm$0.0004 & 0.8789 $\pm$0.0008 & 0.8773 $\pm$0.0003 \\
        \bottomrule
    \end{tabular}
\end{table*}

\subsection{Modality Ablation}

Table~\ref{tab: ablation} reports the results of a leave-one-modality-out ablation, considering all four combinations of three input modalities: thigh accelerometer (act), wrist accelerometer (acw), pressure mat (pm), and depth camera (dc). The combination excluding the depth camera (\{act, acw, pm\}) achieves the lowest F1-scores in both centralised (0.8480) and federated (0.7237) training, indicating the importance of depth information for multimodal HAR. Excluding the pressure mat (\{act, acw, dc\}) also leads to a considerable performance reduction, particularly under federated training. The best-performing combination differs across the two settings: \{acw, pm, dc\} achieves the highest centralised F1-score (0.9281), whereas \{act, pm, dc\} performs best under federated training (0.8936). Thus, both settings benefit from retaining the pressure mat and depth camera, while the relative contribution of the two accelerometer modalities varies with the training setting.

\subsection{Computational Cost and Model Complexity}
\label{sec:complexity}

Table~\ref{tab:model_complexity} compares the total and trainable parameters and the training time of the eight architectures. For Time-MAGNET, LoRA reduces the number of trainable parameters from 55.83M to 47.54M (14.85\%) and reduces training time by approximately 6.2\% under centralised training and 7.8\% under federated training. 

Chronos-MAGNET has 65.81M trainable parameters and requires the longest federated training time among the eight models (10,386 s). In contrast, Chronos-T5 has the shortest training time (2,510 s centrally and 6,620 s federally), but its predictive performance is considerably lower. DeepConvLSTM also incurs substantial computational cost, requiring the longest training time in both settings (6,045 s centrally and 12,589 s federally), despite having fewer parameters than Time-MAGNET.

Late Fusion and Attention Fusion have smaller parameter counts (27.94M and 29.16M, respectively) and shorter training times than Time-MAGNET. However, their lower predictive performance, as reported in Table~\ref{tab:all_models}, reflects a trade-off between computational cost and classification performance. Therefore, Time-MAGNET with LoRA offers a favourable balance between predictive performance, parameter efficiency, and training time.

% Beyond predictive accuracy, practical deployment of a federated multimodal model also depends on its parameter budget and training cost. Table~\ref{tab:model_complexity} reports total and trainable parameter counts alongside centralised and federated training time for all eight architectures.

\begin{table*}[h!]
\caption{Model Complexity and Training Time Comparison}
\label{tab:model_complexity}
\centering
\begin{tabular}{lcccc}
    \toprule
    \multirow{2}{*}{\textbf{Model}} & \multicolumn{2}{c}{\textbf{Parameter Count}} & \multicolumn{2}{c}{\textbf{Training Time (s)}} \\
    \cmidrule(lr){2-3} \cmidrule(lr){4-5}
    & Total & Trainable & Centralized & Federated \\
    \midrule
    Time-MAGNET (with LoRA) & 55,927,648 & 47,535,456 & 3485.05 & 8052.36 \\
    \midrule
    Time-MAGNET (no LoRA)   & 55,829,344 & 55,829,344 & 3714.76 & 8729.16 \\
    \midrule
    Chronos-T5              & 76,179,719 & 76,179,719 & 2510.07 & 6620.15 \\
    \midrule
    Chronos-MAGNET          & 65,813,344 & 65,813,344 & 4517.16 & 10386.09 \\
    \midrule
    BiLSTM                  & 40,702,983 & 40,702,983 & 4283.07 & 8898.17 \\
    \midrule
    DeepConvLSTM            & 50,038,791 & 50,038,791 & 6045.18 & 12589.18 \\
    \midrule
    Late Fusion             & 27,943,241 & 19,551,049 & 3209.34 & 6994.32 \\
    \midrule
    Attention Fusion        & 29,162,313 & 20,770,121 & 3225.07 & 7120.61 \\
    \bottomrule
\end{tabular}
\end{table*}

\section{Conclusion and Future Work}

In this work, we introduced a novel multimodal HAR framework designed for a decentralized environment. At its core, our system is built upon a customized T5 encoder only architecture, which is adapted for multivariate time series representation using time stamp wise patch based tokenization strategy. We also built DART-CNN for generating highly discriminative spatio temporal embeddings from the image modalities. To effectively capture inter modality relationships, we proposed a robust multimodal fusion module named MAGNET, which integrates GAT and the MoE mechanism. To address data decentralisation and client heterogeneity, we evaluated the proposed architecture in a federated learning setting using the FedPAW algorithm, which enables collaborative model training while keeping raw data local to each client. By leveraging FedPAW, our framework enables collaborative training that keeps data within each client device. Through extensive evaluations on the MEx dataset, our novel graph attention based fusion mechanism demonstrated superior performance and robustness over baseline models in both centralized and federated settings.

Although our current evaluation is conducted only on the MEx dataset, the extensive nature of these experiments demonstrates that the proposed architecture is robust, adaptable, and well positioned to generalize across multiple multimodal HAR datasets in the future. Furthermore, while FL helps to preserve data privacy by keeping raw data local, information can still be leaked through model updates during adversarial gradient inversion attacks. To actively mitigate this risk, future work will explore integrating rigorous Differential Privacy (DP) mechanisms to further secure client server communications. Additionally, extending this decentralized framework to incorporate emerging time-series foundation models presents a promising direction for future research.

\printcredits

\bibliographystyle{cas-model2-names}

\bibliography{cas-refs}



% \section*{Appendix}

\end{document}